\documentclass[lettersize,journal]{IEEEtran}
\usepackage{amsmath,amsfonts}
\usepackage{algorithmic}
\usepackage{algorithm}
\usepackage{array}
\usepackage[caption=false,font=normalsize,labelfont=sf,textfont=sf]{subfig}
\usepackage{textcomp}
\usepackage{stfloats}
\usepackage{url}
\usepackage{verbatim}
\usepackage{graphicx}
\usepackage{cite}
\usepackage[table,xcdraw]{xcolor}
\usepackage{arydshln}

\usepackage{booktabs}
\usepackage{multirow}
\usepackage{hyperref}

\begin{document}

\title{MGD-SAM2: Multi-view Guided Detail-enhanced Segment Anything Model 2 for High-Resolution Class-agnostic Segmentation}

\author{{Haoran Shen\textsuperscript{*}, Peixian Zhuang\textsuperscript{*}, \emph{Senior Member, IEEE}, Jiahao Kou, Yuxin Zeng, Haoying Xu, Jiangyun Li\textsuperscript{\dag}}
\thanks{\textsuperscript{*} Equal Contribution. \textsuperscript{\dag} Corresponding author.}
\thanks{Haoran Shen, Jiahao Kou, Yuxin Zeng, and Haoying Xu are with the School of Automation and Electrical Engineering, University of Science and Technology Beijing, Beijing 100083, China (e-mail: M202220738@xs.ustb.edu.cn, M202420803@xs.ustb.edu.cn, M202421748@xs.ustb.edu.cn, M202420841@xs.ustb.edu.cn).

Peixian Zhuang and Jiangyun Li are with the Key Laboratory of Knowledge Automation for Industrial Processes, Ministry of Education, the School of Automation and Electrical Engineering, University of Science and Technology Beijing, Beijing 100083, China (e-mail: zhuangpeixian@ustb.edu.cn, leejy@ustb.edu.cn). Jiangyun Li is also with the Shunde Graduate School of University of Science and Technology Beijing, China. (Corresponding author: Jiangyun Li.)

This work was supported in part by the National Natural Science Foundation of China under Grant 62171252, in part by the Fundamental Research Funds for the Central Universities under Grant 00007764.}}

\markboth{Journal of \LaTeX\ Class Files,~Vol.~14, No.~8, August~2021}%
{Shell \MakeLowercase{\textit{et al.}}: A Sample Article Using IEEEtran.cls for IEEE Journals}


\maketitle

\begin{abstract}

Segment Anything Models (SAMs), as vision foundation models, have demonstrated remarkable performance across various image analysis tasks.
Despite their strong generalization capabilities, SAMs encounter challenges in fine-grained detail segmentation for high-resolution class-independent segmentation (HRCS), due to the limitations in the direct processing of high-resolution inputs and low-resolution mask predictions, and the reliance on accurate manual prompts. To address these limitations, we propose MGD-SAM2 which integrates SAM2 with multi-view feature interaction between a global image and local patches to achieve precise segmentation. MGD-SAM2 incorporates the pre-trained SAM2 with four novel modules: the Multi-view Perception Adapter (MPAdapter), the Multi-view Complementary Enhancement Module (MCEM), the Hierarchical Multi-view Interaction Module (HMIM), and the Detail Refinement Module (DRM). Specifically, we first introduce MPAdapter to adapt the SAM2 encoder for enhanced extraction of local details and global semantics in HRCS images. Then, MCEM and HMIM are proposed to further exploit local texture and global context by aggregating multi-view features within and across multi-scales. Finally, DRM is designed to generate gradually restored high-resolution mask predictions, compensating for the loss of fine-grained details resulting from directly upsampling the low-resolution prediction maps. Experimental results demonstrate the superior performance and strong generalization of our model on multiple high-resolution and normal-resolution datasets. Code will be available at \url{https://github.com/sevenshr/MGD-SAM2}.

\end{abstract}


\section{Introduction}
\IEEEPARstart{H}{igh-Resolution} Class-agnostic Segmentation (HRCS) aims to identify and segment category-agnostic foreground objects within natural scenes, with broad applications including image editing\cite{goferman2011context,zhou2018improving}, virtual reality\cite{lee2021assessing,long2020optimal}, and 3D shape reconstruction\cite{zhou2020salient,cong2023multi}. Dichotomous Image Segmentation (DIS) and High-Resolution Salient Object Detection (HRSOD), as representative tasks in HRCS, necessitate fine-grained segmentation details of objects in high-resolution images. Achieving such precise object delineation requires sophisticated algorithms to balance the global receptive field and local detail perception.

\begin{figure}[htbp]
    \centering
    \includegraphics[width=0.45\textwidth]{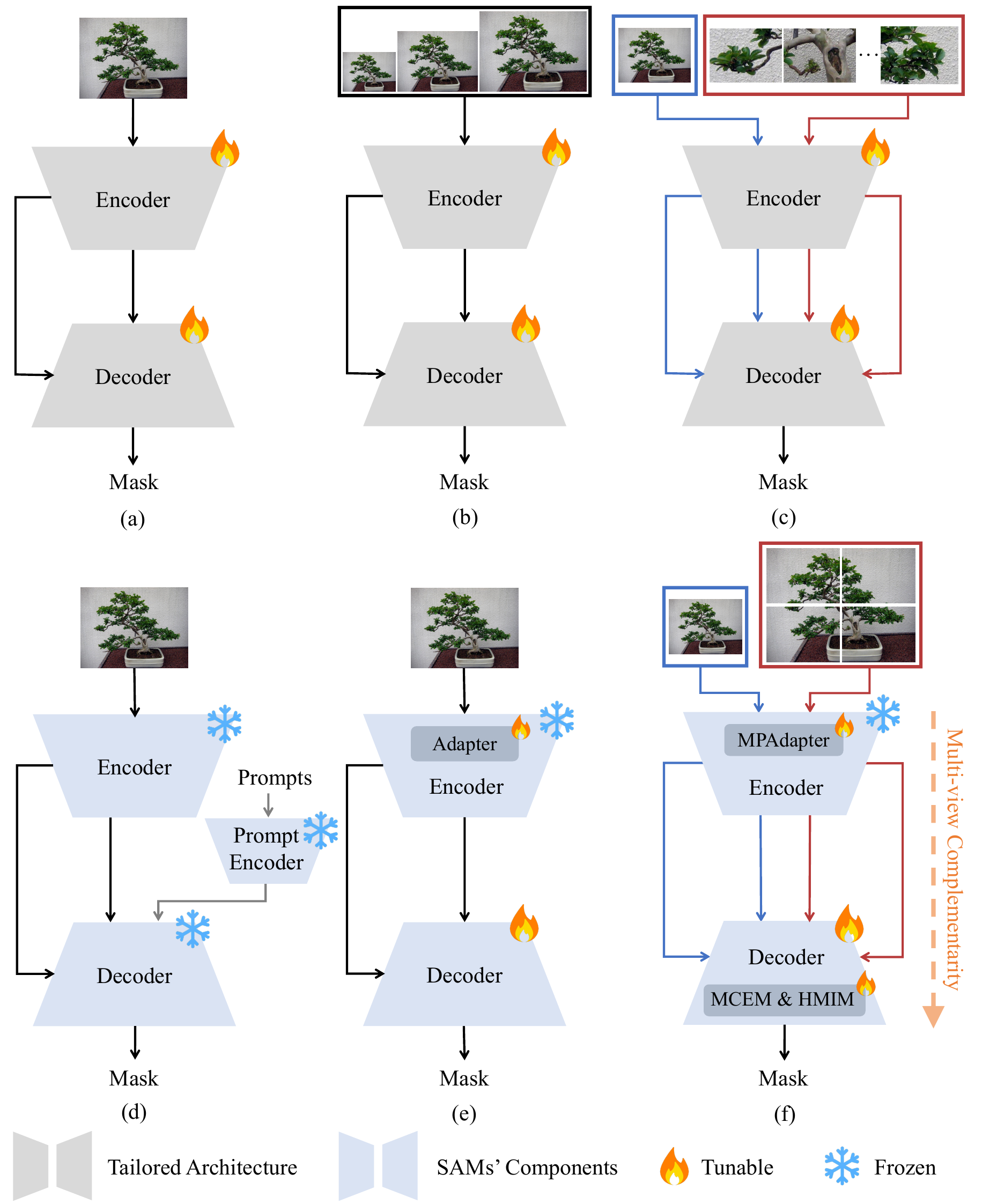}
    \caption{Comparison between the proposed MGD-SAM2 and other existing methods. (a) Common framework\cite{romera2017erfnet,zhang2021looking,wang2022dual}; (b) Image pyramid-based methods\cite{xie2022pyramid,kim2022revisiting}; (c) Patch-based methods\cite{zeng2019towards,yu2024multi}; (d) SAMs' framework\cite{kirillov2023segment,ravi2024sam}; (e) Adapter based SAMs\cite{chen2023sam,gao2024multi,chen2024sam2}; (f) MGD-SAM2: leveraging multi-view complementarity with SAM2's rich prior knowledge for high-resolution class-
    agnostic segmentation.}
    \label{fig_intro}
\vspace{-5pt}
\end{figure}

To solve the above issue, researchers have pursued numerous approaches centered on image scale selection. Common frameworks\cite{romera2017erfnet,zhang2021looking,wang2022dual}, as shown in Fig. \ref{fig_intro}(a), employ a lightweight network to directly process high-resolution inputs, combined with some decoupled local-global perception structures. Image pyramid-based methods\cite{xie2022pyramid,kim2022revisiting}, as depicted in Fig. \ref{fig_intro}(b), employ distinct branches to process images resized at various scales. During the decoding stage, features from larger-scale images are incrementally integrated to enhance detailed information within the global context. Patch-based methods\cite{zeng2019towards,yu2024multi}, as presented in Fig. \ref{fig_intro}(c), refine local details by utilizing image patches instead of the original high-resolution input. The extracted feature regions in the resized raw image are further improved by applying fine-grained patch-wise features. However, common frameworks struggle to balance local-global feature extraction when directly utilizing high-resolution images. Image pyramid-based methods and patch-based methods mitigate the challenges of HRCS images to some extent by processing multi-view images. Nonetheless, these methods integrate features from multi-view images either within a uni-decoder or after individual decoders, lacking explicit multi-view fusion during feature encoding and hierarchical interaction between the multi-stage features of multi-view images. Besides, tailored architecture design mainly focuses on DIS or HRSOD. Therefore, developing a universal model with high precision and robustness for HRCS remains an open challenge.

Recently, the Segment Anything Model (SAM)\cite{kirillov2023segment}, as shown in Fig. \ref{fig_intro}(d), has emerged as a vision foundation model for general image segmentation. Trained on millions of images, SAM exhibits rich, universal image representations and powerful generalization capabilities. The following work, Segment Anything Model 2 (SAM2)\cite{ravi2024sam} further advances accurate and efficient segmentation performance on both image and video tasks. However, despite the general segmentation capabilities of SAMs, there are still limitations when applying them to HRCS\cite{pei2024evaluation}. Firstly, given the demands for fine-grained segmentation, the direct processing of high-resolution images has proven inadequate to capture high-quality global and local features\cite{romera2017erfnet,zhang2021looking}. Secondly, SAMs were originally designed for interactive segmentation, whereas obtaining accurate manual annotations in the HRCS domain is challenging. Finally, the mask results from SAMs's decoder are directly upsampled to yield high-resolution predictions, lacking the reconstruction of fine-grained details. Although many adapter-based methods\cite{chen2023sam,gao2024multi,chen2024sam2}, as depicted in Fig. \ref{fig_intro}(e), have been proposed to overcome the latter two challenges, there is still a lack of introducing local-global representation into SAMs for fine-grained segmentation in HRCS.

To address the aforementioned limitations, we propose MGD-SAM2, as shown in Fig. \ref{fig_intro}(f), which leverages multi-view images to enhance the local-global representation of SAM2. Specifically, we take four non-overlapping sub-images and the resized raw image as multi-view images and send them to the proposed architecture. A Multi-view Perception Adapter (MPAdapter) is designed to improve the ability to perceive global context and local details. Before integrating multi-scale features in SAM2's decoder, the Multi-view Complementary Enhancement Module (MCEM) is proposed to complement the deep features via a local-global cross-attention mechanism. In addition, the Hierarchical Multi-view Interaction Module (HMIM) is devised to utilize multi-scale and multi-view features, further enhancing global semantics and local textures. Finally, by incorporating the mask feature and the unified local image, the Detail Refinement Module (DRM) is proposed to generate higher-resolution masks with finer details.  

In summary, the main contributions of our work are presented as follows:  
\begin{enumerate}
    \item This work is the first attempt to incorporate the general visual prior of SAM2 and the multi-view feature interaction between the resized global image and local patches, enabling fine-grained high-resolution segmentation for universal HRCS task.
    \item We design MPAdapter for specific knowledge injection, MCEM and HMIM for hierarchical feature enhancement, improving global semantics and local texture in feature encoding and decoding by extracting and utilizing hierarchical multi-view features.
    \item We design DRM to generate gradually restored high-resolution masks with finer details by utilizing the mask feature and unified local image, thereby compensating for the detail loss caused by the direct upsampling of low-resolution mask prediction in SAM2. 
    \item Our method outperforms previous SOTA methods and sets new records on both high-resolution datasets (DIS5K, HRSOD, UHRSD, DAVIS-S) and normal-resolution datasets (DUTS, HKU-IS), validating its effectiveness and robustness.

\end{enumerate}

\begin{figure*}[htbp]
    \centering
    \includegraphics[width=1\textwidth]{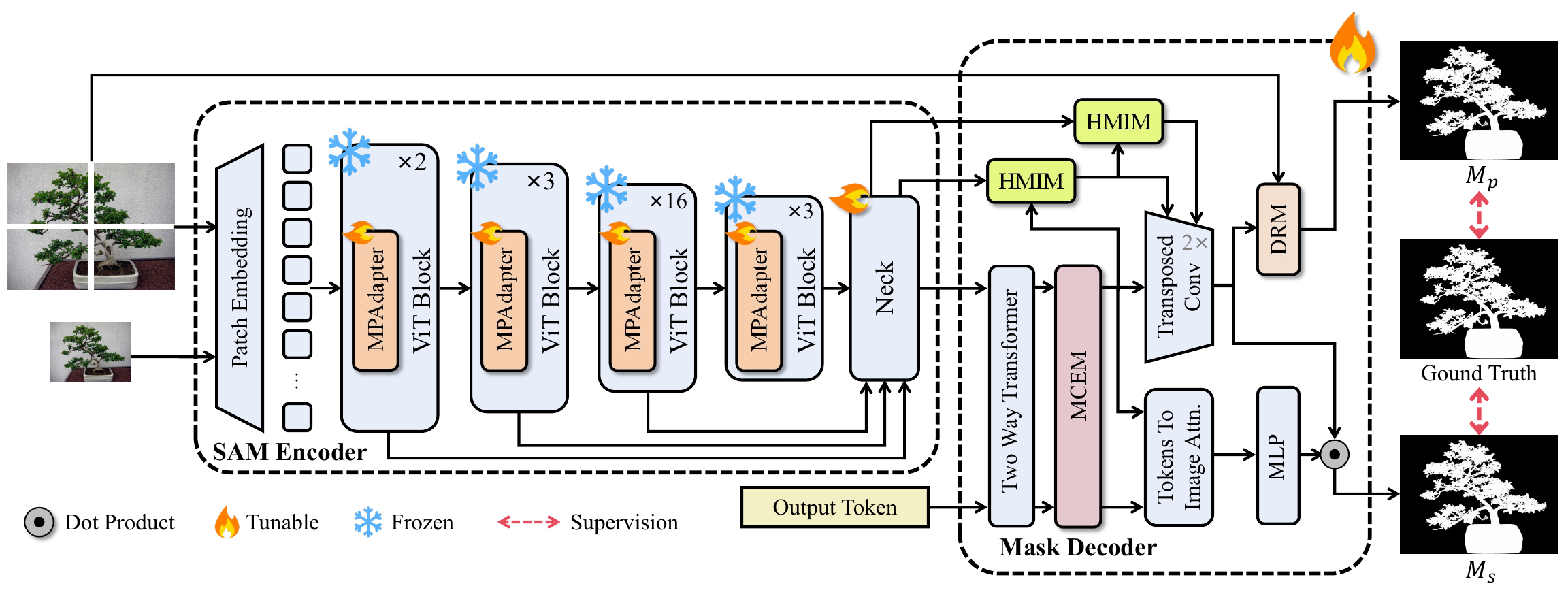}
    \caption{Overall architecture of the proposed MGD-SAM2, which integrates the pre-trained SAM2 with four novel modules: Multi-view Perception Adapter (MPAdapter), Multi-view Complementary Enhancement Module (MCEM), Hierarchical Multi-view Interaction Module (HMIM), and Detail Refinement Module (DRM). We take the combination of image
    patches and the resized raw image as the multi-view input. Firstly, we employ the MPAdapter-assisted SAM2 encoder to extract multi-stage features suitable for the HRCS task. Then, MCEM and  HMIM are proposed to utilize multi-scale and multi-view features, further enhancing the global semantics and local texture. Finally, DRM is proposed to generate the gradually restored high-resolution prediction. Both the upsampled mask prediction $M_s$ from SAM2 decoder and the final mask prediction $M_p$ are supervised. The detailed presentation of each module is shown in Section. \ref{sec_mpadapter}, \ref{sec_mcem}, \ref{sec_hmim}, \ref{sec_drm}.}
    \label{fig_overall}
    \vspace{-5pt}
\end{figure*}

\section{Related Works}
\subsection{High-Resolution Class-agnostic Segmentation}
\subsubsection{High-Resolution Salient Object Detection}
Aiming to identify the most attractive objects in images, Salient Object Detection (SOD) has been studied as a crucial computer vision task for a long time. Numerous approaches \cite{liu2019simple,wei2020f3net,zhao2022position,zhu2023priornet} have been proposed to improve performance on normal-resolution images by aggregating multi-level features. For instance, Liu \textit{et al.}\cite{liu2019simple} proposed a feature aggregation module to integrate multi-scale features adopted from the decoder via pyramid pooling modules. Wei \textit{et al.} \cite{wei2020f3net} proposed to selectively employ multi-level features for better feature complementarity without introducing too much redundancy. Zhao \textit{et al.}\cite{zhao2022position} proposed a global position embedding
attention module to reduce the discrepancy between the multi-view features. Zhu \textit{et al.} \cite{zhu2023priornet} proposed global location prior and local contrast prior to improve the aggregation of multi-level features. Although these methods have shown promising performance on normal-resolution images, directly adopting them for high-resolution images yields unsatisfactory results. This inadequacy arises from their inability to balance receptive field expansion and detail preservation.

Nowadays, with the widespread availability of high-resolution images, there is an increasing demand for high-resolution salient object detection. In response to this trend, Zeng \textit{et al.} \cite{zeng2019towards} contributed the first high-resolution saliency detection dataset, with 1,610 training images and 400 testing images. Besides, they also devised a baseline model by jointly utilizing global guidance and local details. After that, Zhang \textit{et al.} \cite{zhang2021looking} proposed two individual refinement heads to decouple the detail and context information. Similarly, Tang \textit{et al.} \cite{tang2021disentangled} disentangled SOD into a low-resolution saliency classification network for capturing semantics and a high-resolution refinement network for refining details. Besides, Deng \textit{et al.} \cite{deng2023recurrent} proposed RMFormer, which utilizes lower-resolution predictions to guide the generation of high-resolution saliency maps through multi-scale refinement architectures. Recently, Liu \textit{et al.} proposed a two-stage ESNet, with an evolution stage for semantic location and a succession stage for detail refinement. However, the above methods tailored architectures to leverage multi-view and multi-scale features only in feature decoding, limiting their generalization and efficiency in HRCS. In this work, we integrate multi-view interaction with SAM2’s general prior in both the encoding and decoding stages to achieve fine-grained segmentation in universal HRCS.

\subsubsection{Dichotomous Image Segmentation}

DIS is a newly proposed task that emphasizes segmenting class-agnostic objects with varying structural complexities in high-resolution images, making it even more challenging. Qin \textit{et al.} \cite{qin2022highly} first proposed the DIS5K dataset and ISNet with intermediate supervision to preserve fine details. Kim \textit{et al.} \cite{kim2022revisiting} presented an image pyramid-based network named InSPyReNet, enabling the ensemble of multiple results through pyramid-based image blending. Pei \textit{et al.} \cite{pei2023unite} employed a unite-divide-unite framework to disentangle the identification of trunk and structure. Besides, Zhou \textit{et al.} \cite{zhou2023dichotomous} proposed a frequency prior generator and a feature harmonization module to boost the fine-grained segmentation. Similarly, Jiang \textit{et al.} \cite{jiang2024high}  employed a multi-modality fusion module and a collaborative scale fusion module to identify fine-grained object boundaries by leveraging frequency priors and multi-scale aggregation.
Recently, Yu \textit{et al.} \cite{yu2024multi} proposed MVANet for high-resolution images by aggregating the features extracted from image patches and resized raw images within one encoder-decoder structure. Despite the impressive performance of the aforementioned methods, there remains room for improvement in extracting and leveraging multi-view, multi-scale features. More importantly, tailored network designs often exhibit poor generalization capabilities for universal HRCS. Our work leverages exceptional feature extraction and generalization abilities of SAM2, and adapt it to HRCS for superior performance.

\subsection{Segment Anything Model}

SAM \cite{kirillov2023segment} is an interactive segmentation model capable of generating class-agnostic masks based on various prompts. Trained on an extensive dataset with 11 million images, SAM has exhibited remarkable zero-shot transfer capability in numerous natural scenarios. Great efforts \cite{ke2023segment,zhang2024efficientvit,xiong2024efficientsam} have been conducted to improve the segmentation accuracy and computational efficiency of SAM. Moreover, SAM has found widespread applications in various fields, including referring segmentation\cite{lai2024lisa}, video tracking \cite{cheng2023tracking} and medical image processing \cite{ma2024segment}. On top of SAM, SAM2 further extends its capability to video task, enabling effective promptable segmentation across both images and videos. Beyond leveraging the zero-shot abilities, several studies have leveraged SAMs for non-interactive segmentation using parameter-efficient fine-tuning. For instance, Wu \textit{et al.} \cite{wu2023medical} explored several adapters to enhance SAM's segmentation capability for 2D and 3D medical images. Chen \textit{et al.} \cite{chen2023sam,chen2024sam2} incorporated task-specific knowledge into SAMs via adapters for challenging tasks such as shadow detection and camouflaged object detection. Guo \textit{et al.} \cite{gao2024multi} proposed MDSAM for SOD, which leverages multi-scale and multi-level information with lightweight module injections. However, since the above fine-tuning methods are designed for normal-resolution images, they struggle to produce satisfactory results on HRCS that requires fine-grained detail segmentation. In this work, we integrate multi-view complementarity in both the encoding and decoding stages to achieve global localization and local refinement.

\section{Methodology}
\begin{figure}[t]
    \centering
    \includegraphics[width=0.48\textwidth]{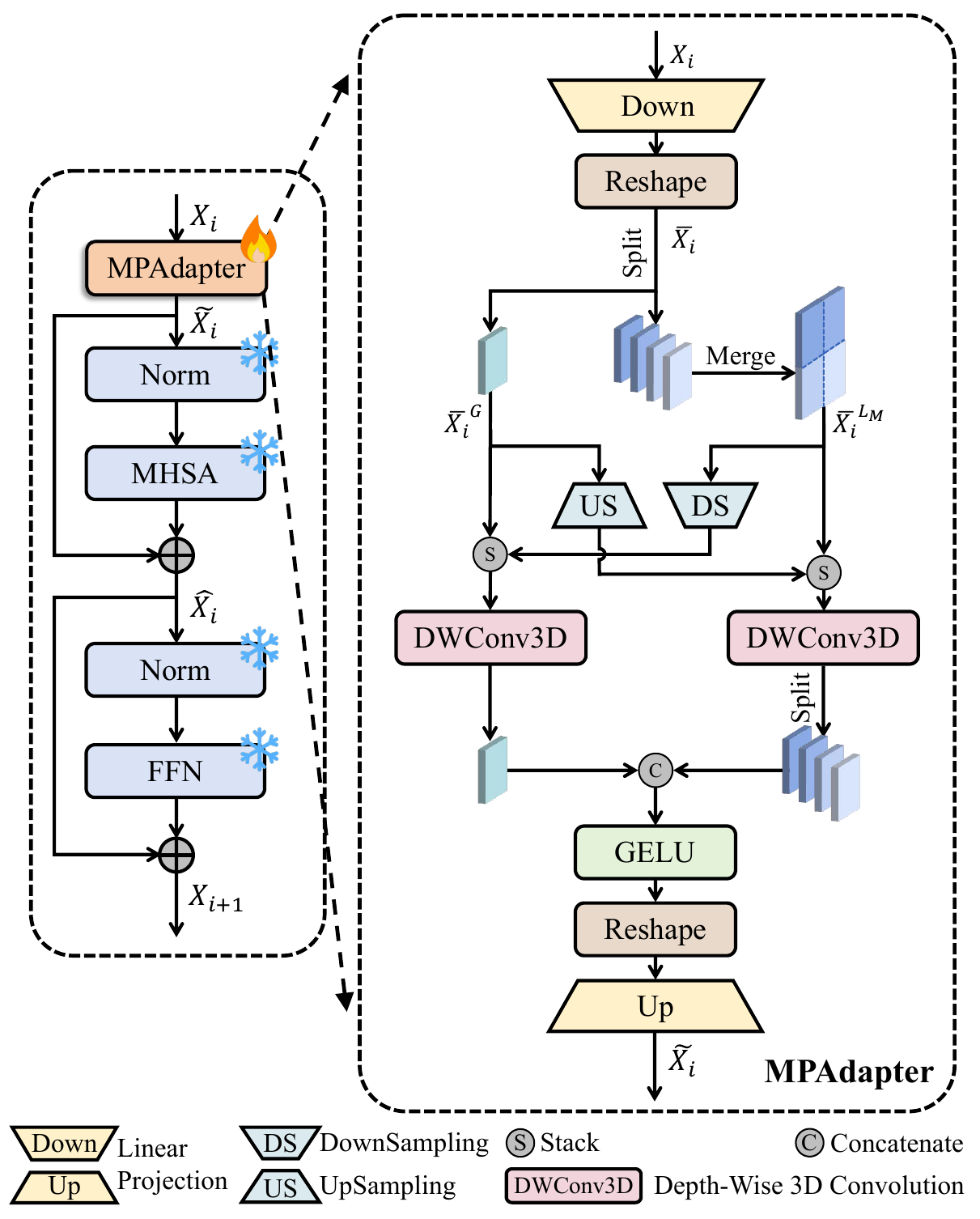}
    \caption{Details of the MPAdapter. To adapt SAM2 to HRCS task in an effective and efficient way, we insert the trainable MPAdapter into each Transformer block of the SAM2's frozen encoder. Based on the Adapter, MPAdapter utilizes two lightweight 3D depth-wise convolutions to enhance local and global features respectively.}
    \label{fig_adapter}
    \vspace{-5pt}
\end{figure}
In this work, a novel MGD-SAM2 is proposed for the HRCS task. As shown in Fig. \ref{fig_overall}, we take the combination of image patches and the resized raw image as input, and send the input to the modified SAM2 with four novel modules: Multi-view Perception Adapter (MPAdapter), Multi-view Complementary Enhancement Module (MCEM), Hierarchical Multi-view Interaction Module (HMIM) and Detail Refinement Module (DRM). Specifically, the original image $I \in \mathbb{R}^{B \times 3 \times H \times W}$ is resized to generate the global image $G \in \mathbb{R}^{B \times 3 \times h \times w}$. And we split the raw image to get four non-overlapping local images $\{L_m\}_{m=1}^4 \in \mathbb{R}^{B \times 3 \times h \times w}$. Then, the combination of global and local images $\{L, G\}\in \mathbb{R}^{5B \times 3 \times h \times w}$ is set as input to our MGD-SAM2. The input is first sent to the SAM2's encoder inserted with MPAdapter to get the hierarchical features $\{F_4,F_8,F_{16}\}$. After the Two Way Transformer, the deep feature $F_{16}$ is further processed through MCEM to obtain more precise local and global localization. Before directly supplementing the shallow features $\{F_4,F_8\}$ to obtain mask prediction of SAM2 decoder, we employ HMIM to utilize shallower merged local features and deeper global features to further enhance the representation of local detail and global context. Finally, DRM utilizes the mask feature and unified local image to obtain progressively restored high-resolution prediction results. A detailed description of each component is provided in the following sections. Table \ref{definition} presents the definition of the main notations used in our MGD-SAM2.

\subsection{Multi-view Perception Adapter}
\label{sec_mpadapter}
\begin{figure}[t]
    \centering
    \includegraphics[width=0.5\textwidth]{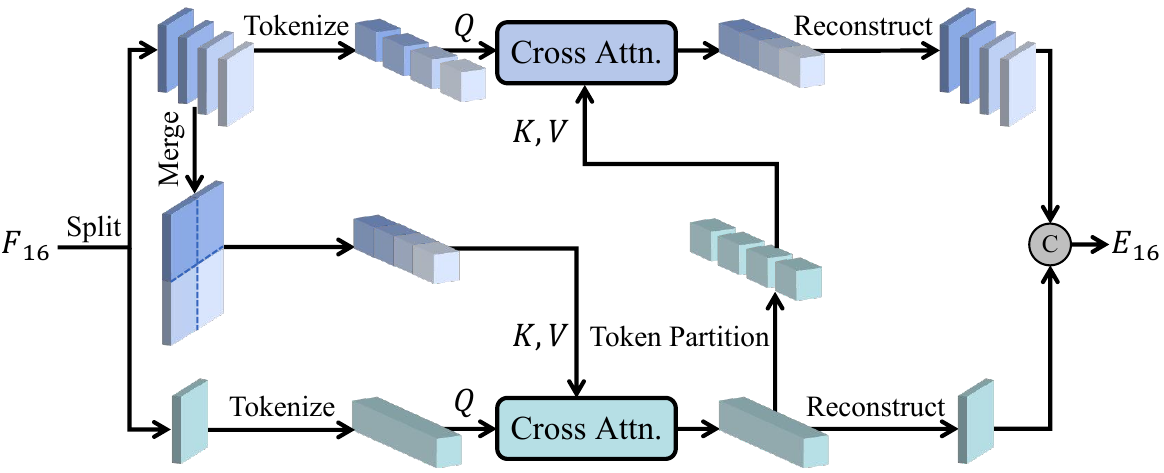}
    \caption{Illustration of the MCEM. After the Two Way Transformer, we employ MCEM to further refine the local and global localization of deep feature $F_{16}$, which utilizes cross-attention to establish the long-range connections between multi-view features.}
    \label{fig_MCEM}
    \vspace{-5pt}
\end{figure}
\begin{table}[t] 
    \caption{Definition of the main notations used in MGD-SAM2.}
    \label{definition}
    \centering
\resizebox{0.5\textwidth}{!}{  
\setlength{\arrayrulewidth}{0.4pt}
\renewcommand{\arraystretch}{1.3}
    \begin{tabular}{cc} 
    \toprule[2pt]
         \emph{Parameters}&  \emph{Definition}\\  \midrule
            $I$&  The original image\\
            $G$&  The global image\\
            $\{L_m\}_{m=1}^4$& The local images  \\
            $L_M$ &  The unified local image\\ 
            $\{F_4,F_8,F_{16}\}$ & The hierarchical features from the encoder\\
            $\{E_4,E_8,E_{16}\}$& The enhanced hierarchical features from the HCEM and HMIM\\
            $X^L$& The local features of $X$ \\
            $X^{L_M}$& The unified local feature of $X$\\
            $X^G $& The global feature of $X$\\
            $M_p$& The MGD-SAM2’s final output\\
            $M_s$& The auxiliary output from SAM2's decoder  \\
    \bottomrule[2pt]
    \end{tabular}
    }
\end{table}
Instead of full fine-tuning SAM2 for non-interactive segmentation in downstream tasks, we propose a lightweight MPAdapter to adapt SAM2 to HRCS. By leveraging the multi-view features, MPAdapter enhances the perception of local details and global contexts with few training parameters. In this way, our MGD-SAM2 preserves the general visual priors of SAM2, while improving the fine-grained segmentation capability for HRCS.

As illustrated in Fig. \ref{fig_adapter}, each Transformer layer of the SAM2's encoder consists of a Multi-Head Self-Attention (MHSA), a Feed Forward Network (FFN) and two Layer Normalization layers (LN). And we insert the proposed MPAdapter before the pre-normalization in each Transformer layer. Thus, the complete process is expressed as follows:
\begin{align}
\widetilde{X}_{i} &= MDAdapter({X}_{i}),\\
\hat{X}_{i} &= MHSA(LN(\widetilde{X}_{i}))+\widetilde{X}_{i},\\
{X}_{i+1} &= FFN(LN(\hat{X}_{i}))+\hat{X}_{i},
\end{align}
where ${X}_{i} \in \mathbb{R}^{5B \times N \times D}$ is the output of $i$-th Transformer layer, $\widetilde{X}_{i}$, $\hat{X}_{i} \in \mathbb{R}^{5B \times N \times D}$ denote two intermediate outputs. $N$ represents the number of tokens. $D$ is the embedding dimension. 

The detailed structure of our MDAdapter is shown on the right side of Fig. \ref{fig_adapter}.  Concretely, we first employ a linear projection layer to reduce the feature dimension:
\begin{align}
\bar{X}_{i} = \tau({X}_{i}W_i^{down}),
\end{align}
where $W_i^{down} \in \mathbb{R}^{D \times \frac{D}{r}}$ is the trainable parameter of the linear layer, $r$ is the reduction factor. After the reshape operation $\tau(\cdot)$, we get the feature $\bar{X}_{i}$ of shape $\mathbb{R}^{5B \times \frac{D}{r} \times \frac{h}{s} \times \frac{w}{s}}$, where $s \in \{4,8,16,32\}$ denote the downsampling stride. 

Then, we propose a dual complementary mechanism to improve the perception of local details and global semantics. $\bar{X}_{i}$ is first split along the batch dimension to get the four local features $\{\bar{X}_{i}^{L_m}\}_{m=1}^4\in \mathbb{R}^{B \times \frac{D}{r} \times \frac{h}{s} \times \frac{w}{s}}$ and a global feature $\bar{X}_{i}^G\in \mathbb{R}^{B \times 3 \times \frac{h}{s} \times \frac{w}{s}}$. And the four local features are merged to generate the unified local feature $\bar{X}_{i}^{L_M}\in \mathbb{R}^{B \times \frac{D}{r} \times \frac{2h}{s} \times \frac{2w}{s}}$ according to their positions in the original image:
\begin{align}
\bar{X}_{i}^{L_M} = Merge(\bar{X}_{i}^{L}),\{\bar{X}_{i}^{L},\bar{X}_{i}^G\} = Split(\bar{X}_{i}).
\end{align}
After unifying the spatial dimension and stacking, we employ two 3D depth-wise convolutions with a stride of $(2, 3, 3)$ to enhance the global features and the unified local features separately:
\begin{align}
\bar{X}_{i}^{L_M} = DWConv3D(Stack(\bar{X}_{i}^{L_M},US(\bar{X}_{i}^G))),\\
\bar{X}_{i}^{G} = DWConv3D(Stack(DS(\bar{X}_{i}^{L_M}),\bar{X}_{i}^G)),
\end{align}
where $US$ and $DS$ are the downsampling and upsampling operations. Afterwards, we split the unified local feature along the spatial dimension and concat them with the global feature to get $\bar{X}_{i} \in \mathbb{R}^{5B \times \frac{D}{r} \times 1 \times \frac{h}{s} \times \frac{w}{s}}$:
\begin{align}
\bar{X}_{i} = Concat(Split(\bar{X}_{i}^{L_M}),\bar{X}_{i}^{G}).
\end{align}
Finally, we reshape the enhanced feature $\bar{X}_{i}$ activated by the GELU function to the size of $5B \times N \times \frac{D}{r}$. With a linear projection layer $W_i^{up} \in \mathbb{R}^{\frac{D}{r} \times D}$  to restore the feature dimension, we obtain the output of MPAdapter:
\begin{align}
\widetilde{X}_{i} = \tau(GELU(\bar{X}_{i}))W_i^{up}.
\end{align}

\begin{figure}[!t]
    \centering
    \includegraphics[width=0.5\textwidth]{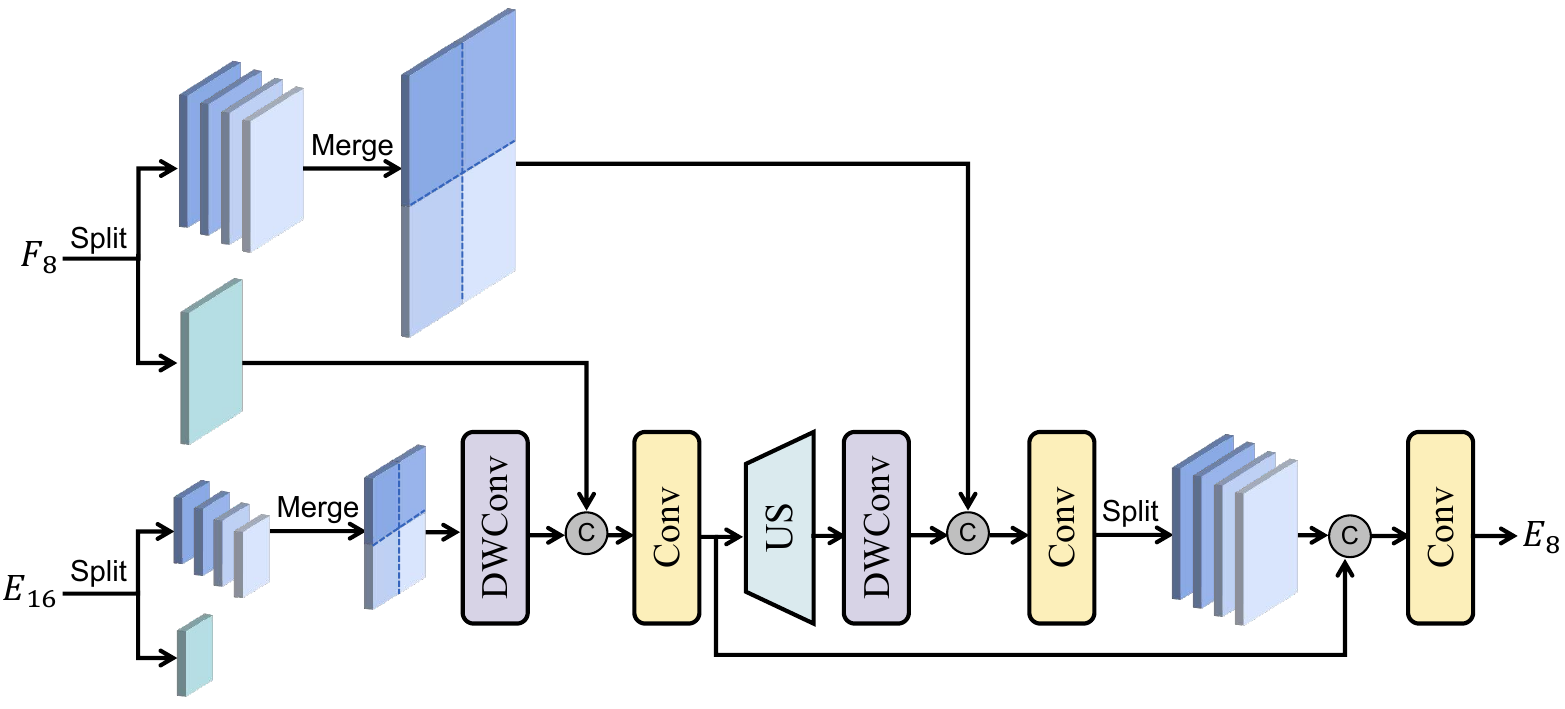}
    \caption{Illustration of the HMIM. Before directly using the deep feature $E_{16}$ and shallow features $\{F_4,F_8\}$ to obtain SAM2's mask prediction, we devise HMIM to leverage multi-scale and multi-view features to improve the local detail and global context of shallow features.}
    \label{fig_HMIM}
    \vspace{-5pt}
\end{figure}

\subsection{Multi-view Complementary Enhancement Module}
\label{sec_mcem}
After the Two Way Transformer, we employ MCEM to improve the local and global localization of deep feature $F_{16}$ via a cross-attention mechanism. As shown in Fig. \ref{fig_MCEM}, we first split and merge the input feature to get the global feature $F_{16}^{G} \in \mathbb{R}^{B \times C \times \frac{h}{16} \times \frac{w}{16}}$ and unified local feature $F_{16}^{L_M} \in \mathbb{R}^{B \times C \times \frac{h}{8} \times \frac{w}{8}}$. Then, the local and global features are tokenized to get $F_{T}^{L_M}$ and $F_{T}^{G}$. Afterwards, we take $F_{T}^{G}$ as $Q$, $F_{T}^{L_M}$ as $K$, $V$ and send them to the Multi-Head Cross-Attention
(MHCA), followed by LN and FFN, which is abbreviated as "Cross Attn." in green in Fig. \ref{fig_MCEM}:
\begin{align}
\hat{F}_{T}^{G} &= LN(MHCA(F_{T}^{G},F_{T}^{L_M},F_{T}^{L_M})+F_{T}^{G}),\\
E_{T}^{G} &= LN(FFN(\hat{F}_{T}^{G})+\hat{F}_{T}^{G}).
\end{align}
Subsequently, the enhanced global feature $E_{T}^{G}$ is utilized to assist the refinement of local features. According to the position correlation, we partition $E_{T}^{G}$ into 4 patch tokens $\{E_{T}^{G_m}\}_{m=1}^4\in \mathbb{R}^{B \times \frac{hw}{32^2} \times C}$. Taking the tokenized local features $F_{T}^{L_m}$ as $Q$, $E_{T}^{G_m}$ as $K$, $V$, we employ MHCA within each patch, followed by LN and a shared FFN to generate the enhanced local features $E_{T}^{L_m}$. This operation is abbreviated as "Cross Attn." in blue in Fig. \ref{fig_MCEM}:
\begin{align}
\hat{F}_{T}^{L_m} &= LN(MHCA(F_{T}^{L_m},E_{T}^{G_m},E_{T}^{G_m})+F_{T}^{L_m}),\\
E_{T}^{L_m} &= LN(FFN(\hat{F}_{T}^{L_m})+\hat{F}_{T}^{L_m}).
\end{align}
Finally, the enhanced local and global features are unflattened and reshaped to form the reconstructed features. And they are concatenated in batches to generate the enhanced deep feature $E_{16}$ for subsequent processing.
\begin{figure}[!t]
    \centering
    \includegraphics[width=0.5\textwidth]{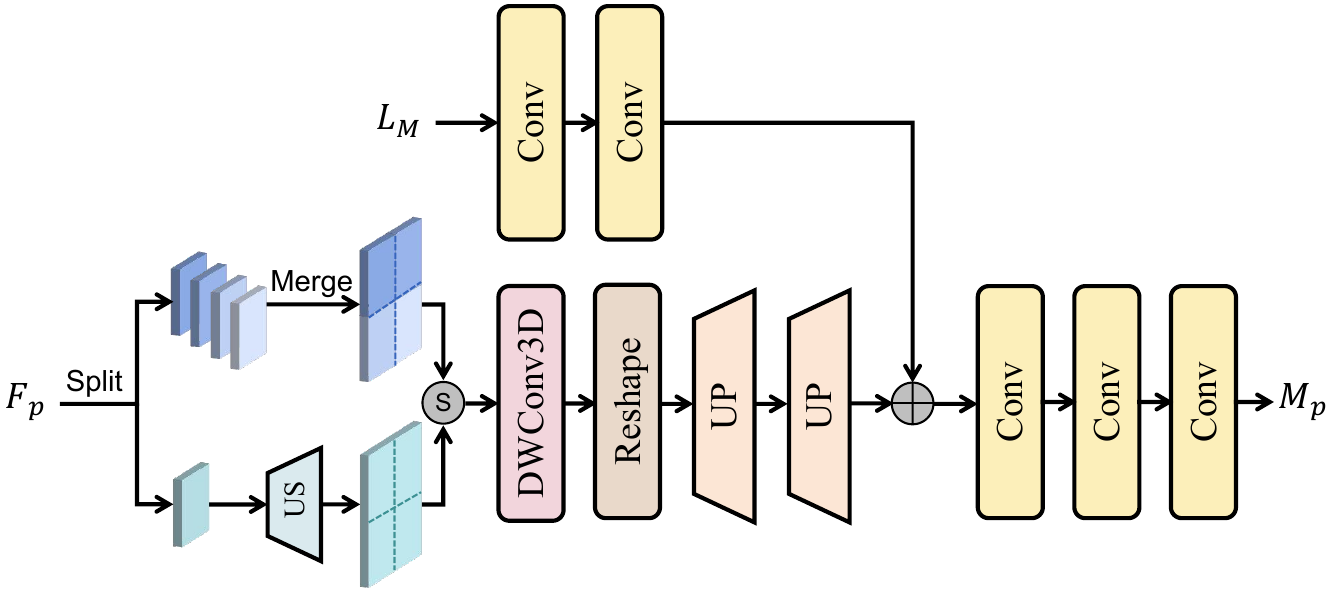}

    \caption{Illustration of the DRM. To compensate for the lack of details in high-resolution predictions directly upsampled by SAM2, we propose DRM to incrementally restore fine-grained details using mask feature $F_p$ and the unified local image $L_M$.}
    \label{fig_DRM}
    \vspace{-5pt}
\end{figure}

\subsection{Hierarchical Multi-view Interaction Module}
\label{sec_hmim}
Instead of directly leveraging $E_{16}$ and the shallow features $\{F_4,F_8\}$ from the encoder to obtain SAM2's mask prediction, we employ HMIM to utilize shallower global features and merged deeper local features to improve the local detail and global context of $\{F_4,F_8\}$. The detailed structure of HMIM is presented in Fig. \ref{fig_HMIM}.

Taking HMIM with inputs $E_{16}$ and $F_8$ as an example, we first generate the unified local feature and global feature through splitting and merging:
\begin{align}
E_{16}^{L_M} = Merge(E_{16}^{L})&,\{E_{16}^{L},E_{16}^G\} = Split(E_{16}),\\
F_{8}^{L_M} = Merge(F_{8}^{L})&,\{F_{8}^{L},F_{8}^G\} = Split(F_{8}),
\end{align}
where $E_{16}^{L_M}$, $F_{8}^L$ and $F_{8}^G$ are of size $B \times C \times \frac{h}{8} \times \frac{w}{8}$, $F_{8}^{L_M}$ is of size $B \times C \times \frac{h}{4} \times \frac{w}{4}$. After a $3 \times3$ depth-wise convolution, $E_{16}^{L_M}$ is 
concatenated with $F_{8}^G$ in channel dimension. Then the enhanced global feature $\hat{E}_{8}^G$ is obtained through a $1 \times1$ convolution. Subsequently, the upsampled $\hat{E}_{8}^G$ and $F_{8}^{L_M}$ go through another combination of a $3 \times3$ depth-wise convolution and a $1 \times1$ convolution to get the enhanced unified local feature $\hat{E}_{8}^{L_M}$. Finally, we split $\hat{E}_{8}^{L_M}$ in spatial dimension and concat it with $\hat{E}_{8}^G$. With the combination of a layer normalization, a GELU function, and a $1 \times1$ convolution, we obtain the enhanced shallow feature ${E}_{8}$.

Similarly, we improve the local and global representation of ${F}_{4}$ through HMIM to get ${E}_{4}$. Subsequently, ${E}_{4}$, ${E}_{8}$ and ${E}_{16}$ are fed into the remaining part of SAM2 decoder to generate mask prediction $M_s$.

\subsection{Detail Refinement Module}
\label{sec_drm}
With the assistance of MPAdpter, MCEM, and HMIM, our method sufficiently leverages the multi-view and multi-scale features to adapt SAM2 for HRCS. However, the mask prediction upsampled in SAM2 leads to insufficient detail restoration. To address this issue, we propose Detail Refinement Module (DRM) to utilize the mask feature and unified local image to generate high-resolution prediction results with fine-grained details.

As shown in Fig. \ref{fig_DRM}, DRM consists of a main branch and an auxiliary branch. The main branch gradually restores the mask feature to the output resolution, while the auxiliary branch incorporates the detailed feature extracted from the unified local image. Specifically, in the main branch, we first split and merge the mask feature $F_p$ to get the global feature $F_{p}^{G}$ and unified local feature $F_{p}^{L_M}$. Then, we employ a 3D depth-wise convolution to fuse the stacked ${F}_{p}^{L_M}$ and upsampled ${F}_{p}^G$. Afterwards, we progressively restore the feature to the input resolution by bilinear interpolations and 3 × 3 convolutions. In the auxiliary branch, we use some $3\times3$ convolutional layers to get the unified local image feature. Finally, we add the restored
feature and the unified local image feature, and employ another combination of convolutions to get the fine-grained mask prediction $M_p$.
\begin{figure*}[htbp]
    \centering
    \includegraphics[width=\textwidth]{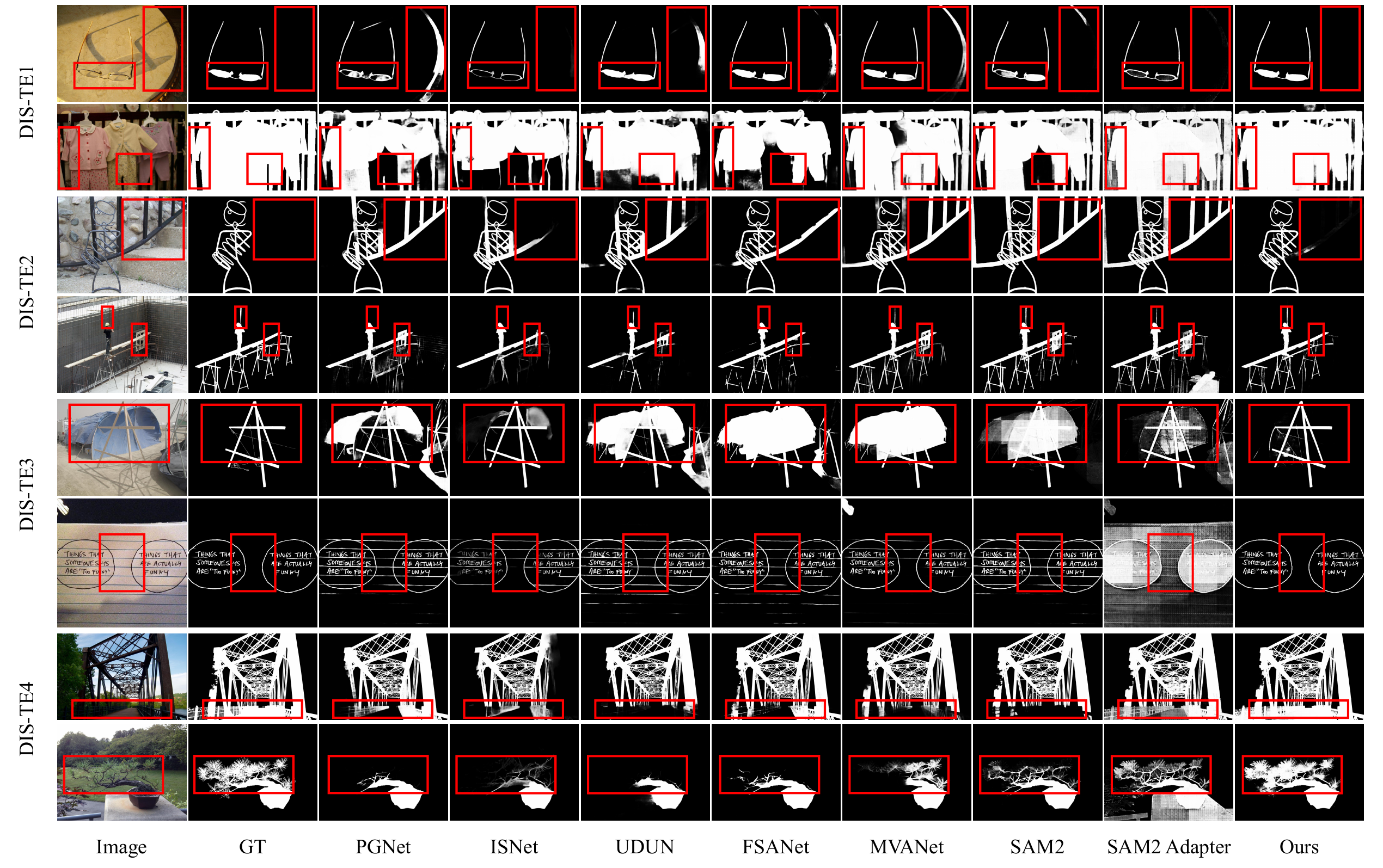}
    \caption{Qualitative comparison between the proposed MGD-SAM2 and other existing methods on DIS5K, including PGNet\cite{xie2022pyramid}, ISNet\cite{qin2022highly}, UDUN\cite{pei2023unite}, FSANet\cite{jiang2024high}, MVANet\cite{yu2024multi}, SAM2\cite{ravi2024sam}, and SAM2 Adapter\cite{chen2024sam2}.}
    \label{visual_dis}
\end{figure*}

\subsection{Loss Function}
\begin{figure*}[htbp]
    \centering
    \includegraphics[width=\textwidth]{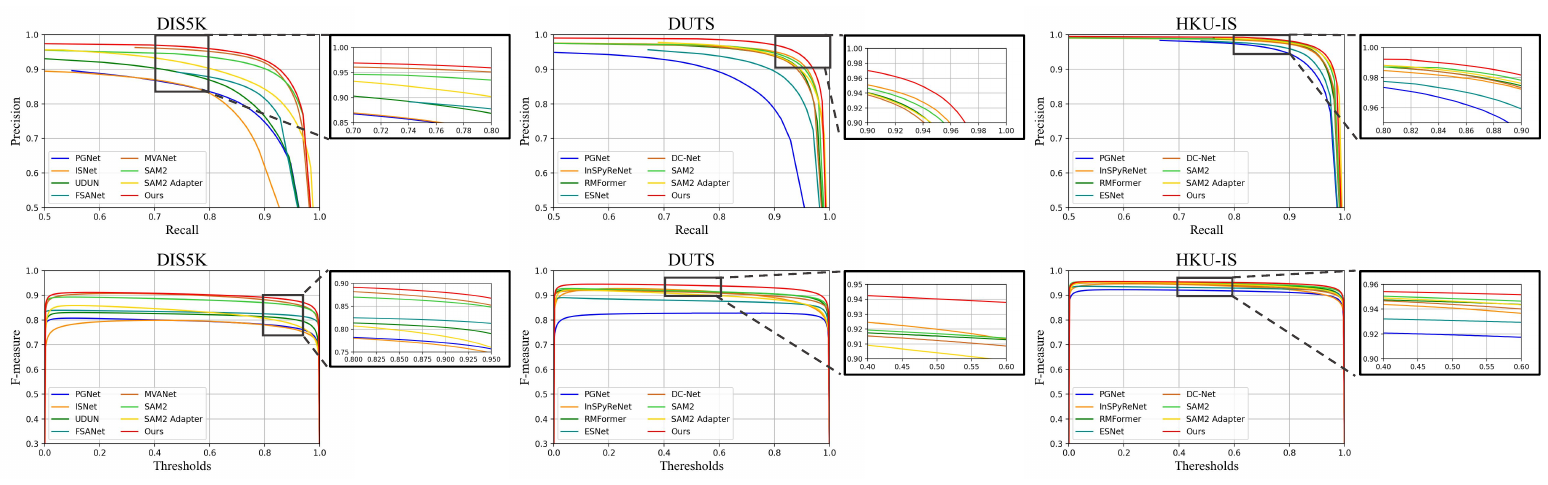}
    \caption{PR and F-measure curves of the proposed model and other existing methods on DIS5K, DUTS, and HKU-IS, including PGNet\cite{xie2022pyramid}, ISNet\cite{qin2022highly}, 
    InSPyReNet\cite{kim2022revisiting}, UDUN\cite{pei2023unite}, FSANet\cite{jiang2024high}, MVANet\cite{yu2024multi}, ESNet\cite{liu2024esnet}, DC-Net\cite{zhu2025dc}, SAM2\cite{ravi2024sam}, and SAM2 Adapter\cite{chen2024sam2}.}
    \label{curve_dis}
\end{figure*}

Following previous works \cite{pei2023unite,zhou2023dichotomous}, we use the combination of the binary cross-entropy (BCE) loss and the weighted Interaction over Union (IoU) loss for model training:
\begin{align}
\mathcal{L}(M,M_{gt}) = \mathcal{L}_{BCE} + \mathcal{L}_{IoU}.
\end{align}
In addition to our MGD-SAM2’s final output $M_p$, we incorporate the mask prediction $M_s$ from SAM2 decoder as auxiliary supervision. Consequently, the total loss is formulated as:
\begin{align}
\mathcal{L}_{total} = \mathcal{L}(M_p,M_{gt}) + \lambda\mathcal{L}(M_s,M_{gt}).
\end{align}
where $\lambda$ is set to 0.3 in our experiment.

\section{Experiments}
In this section, we present the dataset settings, evaluation metrics, implementation details, state-of-the-art comparisons, ablation experiments, and zero-shot analysis. The details are described below.

\begin{figure*}[!t]
    \centering
    \includegraphics[width=\textwidth]{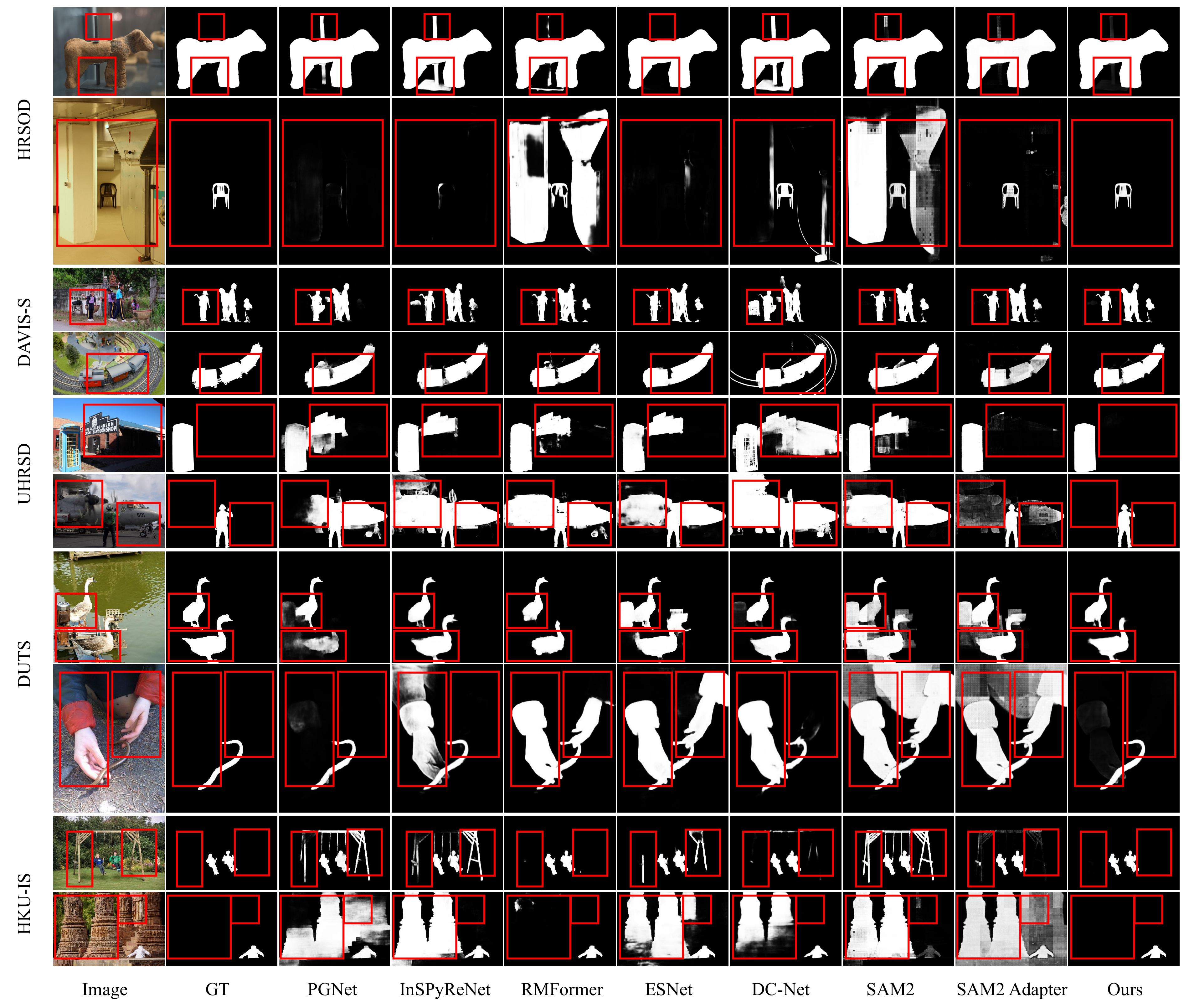}
    \caption{Qualitative comparison between the proposed MGD-SAM2 and other existing methods on HRSOD and SOD datasets, including PGNet\cite{xie2022pyramid}, 
    InSPyReNet\cite{kim2022revisiting}, RMFormer\cite{deng2023recurrent}, ESNet\cite{liu2024esnet}, DC-Net\cite{zhu2025dc}, SAM2\cite{ravi2024sam}, and SAM2 Adapter\cite{chen2024sam2}.}
    \label{visual_sod}
\end{figure*}

\subsection{Datasets}
To evaluate the performance of the proposed MGD-SAM2, we conduct extensive experiments on four high-resolution datasets (DIS5K\cite{qin2022highly} for DIS, HRSOD\cite{zeng2019towards}, DAVIS-S\cite{zeng2019towards} and UHRSD\cite{xie2022pyramid} for HRSOD) and two normal-resolution datasets (DUTS\cite{wang2017learning} and HKU-IS\cite{li2015visual} for SOD). 
\subsubsection{Training Sets}
For DIS, we follow \cite{qin2022highly, pei2024evaluation,zhou2023dichotomous} to use DIS5K-TR as our training set in experiments, which contains 3,000 images. For HRSOD and SOD, we follow \cite{xie2022pyramid} to use the combination of DUTS-TR and HRSOD-TR as the training set, consisting of 10,533 and 1,610 images, respectively.
\subsubsection{Test Sets}
For DIS, we conduct experiments on all test sets in DIS5K (DIS-TE1, DIS-TE2, DIS-TE3, and DIS-TE4), with increasing structure complexities, each containing 500 images. For HRSOD, we evaluate MGD-SAM2 on HRSOD-TE, DAVIS-S, and UHRSD-TE, which include 400, 92, and 988 images, respectively. To assess the generalization ability of our method, we also conduct experiments on two SOD datasets: DUTS-TE and HKU-IS, which contain 5,019 and 4,447 images. Besides, DIS-VD with 470 images is used for ablation study.

\subsection{Evaluation Metrics}

Following previous works, we adopt some widely used metrics to evaluate the performance, including max F-measure ($F_{\beta}^{max}$) \cite{perazzi2012saliency}, weighted F-measure ($F^{\omega}_{\beta}$) \cite{margolin2014evaluate}, mean enhanced-alignment measeure ($E^m_{\phi}$) \cite{fan2018enhanced}, structural measure ($S_m$) \cite{fan2017structure}, and mean absolute error ($\mathcal{M}$) \cite{perazzi2012saliency}. $F_{\beta}^{max}$ and $F^{\omega}_{\beta}$ are the maximum and harmonic mean of precision and recall, respectively, where $\beta$ is set to 0.3. $S_m$ reflects the object-aware and region-aware structural similarity between prediction maps and true maps. $E^m_{\phi}$ measures the pixel-level similarity between prediction and ground truth. $\mathcal{M}$ measures the average error of the predicted and ground truth mask.

\subsection{Implementation Details}

Experiments are implemented in PyTorch on a single RTX 3090 GPU. To prevent overfitting, data augmentation techniques such as random horizontal flipping, cropping, and rotation are applied. The original image is first resized to $1024 \times 1024$. Then, we take the combination of four split local images with a size of $512 \times 512$ and a resized global image of $512 \times 512$ as input. For initialization, we load the pre-trained weights of the image encoder and mask decoder from SAM2-B, while all other parameters of our MGD-SAM2 are initialized randomly. For hyperparameter settings, we use the AdamW optimizer during training. we freeze SAM2's encoder and set the initial learning rate for randomly initialized parameters to $2e^{-4}$, while the learning rate for other pre-trained parameters is reduced by $10\times$. Besides, we train the model for 80 epochs using a batch size of 2.

\subsection{State-of-the-art Comparison}
\begin{table*}[htbp]
\caption{Quantitative results on DIS5K\cite{qin2022highly}. The best result is marked in \textcolor{red}{red}, while the second-best result is marked in \textcolor{green}{green}.}
    \label{tabel_dis}
  \centering
  \resizebox{\textwidth}{!}{  
  \footnotesize
  \renewcommand{\arraystretch}{1.3}
\begin{tabular}{cc|ccccccccccccc|cc|c}
\toprule[2pt]
\multirow{4}{*}{\emph{Datasets}}              & \multirow{2}{*}{\emph{Method}}                        & F\(^3\)Net & GCPANet   & PFNet       & ISDNet    & IFA     &  PGNet    & ISNet    &C2FNet &BCMNet & UDUN   &FP-DIS& FSANet & MVANet   & SAM2    & SAM2 Adapter& \multirow{4}{*}{\emph{Ours}}
\\ 
        &                       & \cite{wei2020f3net} & \cite{chen2020global}    & \cite{mei2021camouflaged}          & \cite{guo2022isdnet}     & \cite{hu2022learning}    &   \cite{xie2022pyramid}      & \cite{qin2022highly}      &\cite{chen2022camouflaged} & \cite{cheng2023bidirectional} &  \cite{pei2023unite}       &  \cite{zhou2023dichotomous}   & \cite{jiang2024high} & \cite{yu2024multi}  &\cite{ravi2024sam}   & \cite{chen2024sam2}  &    \\ \cline{3-17}
      
    &  \multirow{1}{*}{\emph{Publications}}    & AAAI & AAAI & CVPR & CVPR & ECCV & CVPR & ECCV & TCSVT & TCSVT & MM & IJCAI &TNNLS & CVPR  & ICLR & --    \\ \cline{3-17}
    & \multirow{1}{*}{\emph{Years}} & 2020 & 2020 & 2021  & 2022 & 2022 & 2022 & 2022 & 2022& 2023 & 2023 & 2023 &2024 & 2024 & 2025 & 2024   \\ \hline
                                    
    \multirow{5}{*}{\emph{DIS-TE1}}& \(F_{\beta}^{max}\) \(\uparrow\) & 0.726      & 0.741      & 0.740       &            0.717&            0.673&  0.754&             0.740&0.713 &0.714&            0.784&0.784&            0.777 & \textcolor{green}{0.893} & 0.884 & 0.824 & \textcolor{red}{0.914}      \\ 
     & \(F^{\omega}_{\beta}\) \(\uparrow\)      & 0.655      & 0.676      & 0.665      &            0.643&            0.573&  0.680&             0.662 &0.639 &0.642 &            0.720&    0.713 &            0.718&\textcolor{green}{0.823} & 0.813 & 0.743 & \textcolor{red}{0.837}      \\ 
    & \(E^m_{\phi}\) \(\uparrow\)      & 0.820      & 0.834      & 0.830     &            0.824&            0.785&  0.848&             0.820&0.822 &0.836 &            0.864&0.860       &            0.860& \textcolor{green}{0.911} & 0.910 & 0.870 & \textcolor{red}{0.920}     \\ 
    & \(S_m\) \(\uparrow\)             & 0.783      & 0.797      & 0.791     &            0.782&            0.746&  0.800&             0.787&0.775 &0.780 &            0.817&0.821 &            0.821&{0.879}& \textcolor{green}{0.883} & 0.852 & \textcolor{red}{0.895}  \\ 
    & $\mathcal{M}$   \(\downarrow\)              & 0.074      & 0.070      & 0.075      &            0.077&            0.088&  0.067&             0.074&0.077 &0.072&            0.059&0.060      &            0.059& \textcolor{green}{0.037} & 0.040 & 0.055 & \textcolor{red}{0.033}   \\ \hline
\multirow{5}{*}{\emph{DIS-TE2}}& \(F_{\beta}^{max}\) \(\uparrow\) & 0.789      & 0.799      & 0.796     &            0.783&            0.758&  0.807&             0.799&0.773 &0.772 &            0.829&0.827 &0.845& \textcolor{green}{0.925}& 0.910 & 0.858 & \textcolor{red}{0.937}      \\ 
 & \(F^{\omega}_{\beta}\) \(\uparrow\)      & 0.719      & 0.741      & 0.729      &            0.714&            0.666&  0.743&             0.728 &0.703 &0.704 &            0.768&   0.767   &0.800&\textcolor{red}{0.874} & 0.852 & 0.787 & \textcolor{green}{0.872}      \\ 
    & \(E^m_{\phi}\) \(\uparrow\)      & 0.860      & 0.874      & 0.866      &            0.865&            0.835&  0.880&             0.858&0.857 &0.868 &            0.886&0.893 &0.905&\textcolor{red}{0.944}& 0.930 & 0.894 & \textcolor{green}{0.938}   \\
    & \(S_m\) \(\uparrow\)             & 0.814      & 0.830      & 0.821     &            0.817&            0.793&  0.833&             0.823&0.807 &0.808 &            0.843&0.845&0.860&\textcolor{green}{0.915} & 0.904 & 0.873 & \textcolor{red}{0.916} \\ 
    & $\mathcal{M}$  \(\downarrow\)              & 0.075      & 0.068      & 0.073    &            0.072&            0.085&             0.065&             0.070&0.075 &0.073 &            0.058&0.059&0.052&\textcolor{red}{0.030} & 0.037 & 0.057 & \textcolor{red}{0.030}      \\ \hline
           \multirow{5}{*}{\emph{DIS-TE3}}                         & \(F_{\beta}^{max}\) \(\uparrow\) & 0.824      & 0.844      & 0.835      &            0.817&            0.797&             0.843&             0.830&0.813 &0.802 &            0.865&0.868 &0.870&\textcolor{green}{0.936}& 0.925 & 0.888 & \textcolor{red}{0.947}      \\ 
           & \(F^{\omega}_{\beta}\) \(\uparrow\)      & 0.762      & 0.789      & 0.771     &            0.747&            0.705&  0.785&             0.758 &0.743 &0.734 &            0.809&      0.811&0.827 &\textcolor{green}{0.890}& 0.861 & 0.827 & \textcolor{red}{0.895}     \\ 
        & \(E^m_{\phi}\) \(\uparrow\)      & 0.892      & 0.909      & 0.901     &            0.893&            0.861&             0.911&             0.883&0.891 &0.897 &            0.917&0.922 &0.923&\textcolor{red}{0.954}&0.939 &0.923 & \textcolor{green}{0.950}     \\ 
          & \(S_m\) \(\uparrow\)             & 0.841      & 0.855      & 0.847     &            0.834&            0.815&             0.844&             0.836&0.828 &0.823 &            0.865&0.871 &0.873& \textcolor{green}{0.920}&0.910 &0.893 & \textcolor{red}{0.925}      \\
            & $\mathcal{M}$  \(\downarrow\)              & 0.063      & 0.068      & 0.062     &            0.065&            0.077&             0.056&             0.064&0.068 &0.069 &            0.050&0.049&0.047 &\textcolor{green}{0.031}& 0.035 & 0.045 & \textcolor{red}{0.028}     \\ \hline
 \multirow{5}{*}{\emph{DIS-TE4}}                       & \(F_{\beta}^{max}\) \(\uparrow\) & 0.815      & 0.831      & 0.816     &            0.794&            0.790&             0.831&             0.827&0.785 &0.776 &            0.846&0.846  &0.872& \textcolor{green}{0.911}& 0.910 & 0.870 & \textcolor{red}{0.932}     \\
 & \(F^{\omega}_{\beta}\) \(\uparrow\)      & 0.753      & 0.776      & 0.755     &            0.725&            0.700&  0.774&             0.753 &0.717 &0.706 &            0.792&    0.788  &0.831&0.857& \textcolor{green}{0.858} & 0.800 & \textcolor{red}{0.877}       \\ 
    & \(E^m_{\phi}\) \(\uparrow\)      & 0.883      & 0.898      & 0.885     &  0.873          & 0.847           &             0.899&             0.870&0.867 &0.876 &            0.901&0.906 &0.926& \textcolor{green}{0.944}& 0.936 & 0.904 & \textcolor{red}{0.948}     \\ 
    & \(S_m\) \(\uparrow\)             & 0.826      & 0.841      & 0.831       &      0.815             &      0.841&       0.811     &             0.830&0.807 &0.803 &            0.849&0.852 &0.874&\textcolor{green}{0.903}& 0.901 & 0.871 & \textcolor{red}{0.913}     \\ 
            & $\mathcal{M}$  \(\downarrow\)              & 0.070      & 0.064      & 0.072           &         0.079   &         0.085  &             0.065 &             0.072&0.082 &0.084 &            0.059&0.061&0.050& \textcolor{green}{0.041} & \textcolor{green}{0.041} & 0.058 & \textcolor{red}{0.034}      \\ \midrule
\multirow{5}{*}{\emph{Overall}}& \(F_{\beta}^{max}\) \(\uparrow\) & 0.789& 0.804& 0.797&0.778 &0.755&  0.809 &  0.799&0.771 &0.766 & 0.831&0.831 &0.841&\textcolor{green}{0.916}& 0.907 & 0.860 & \textcolor{red}{0.933}\\
& \(F^{\omega}_{\beta}\) \(\uparrow\)      & 0.722      & 0.746      & 0.730     &            0.707&            0.661&  0.746&             0.726 &0.701 &0.607 &            0.772&   0.770   &0.794&\textcolor{green}{0.855} & 0.846 & 0.789 & \textcolor{red}{0.870}       \\ 
& \(E^m_{\phi}\) \(\uparrow\)      & 0.864&0.879 &0.871  &0.864 &0.832&  0.885 &  0.858&0.859 &0.869 & 0.892&0.895 &0.903&\textcolor{green}{0.938}& 0.929 & 0.898 & \textcolor{red}{0.939}\\
& \(S_m\) \(\uparrow\)             &0.816 & 0.831& 0.823 & 0.812& 0.791&  0.830& 0.819 &0.804 &0.804 & 0.844&0.847 &0.857&\textcolor{green}{0.905}& 0.899 & 0.897 & \textcolor{red}{0.912}\\
& $\mathcal{M}$   \(\downarrow\)              & 0.071& 0.065& 0.071 & 0.073& 0.084&  0.063&  0.070&0.076 &0.075& 0.057&0.057 &0.052&\textcolor{green}{0.035}& 0.038 & 0.054 & \textcolor{red}{0.031}\\
\bottomrule[2pt]
\\
\end{tabular}
}
\vspace{-5pt}
\label{SOTA_result_dis}
\end{table*}

\begin{table*}[htbp]
\label{table_hrsod}
\caption{Quantitative results on the high-resolution HRSOD\cite{zeng2019towards}, DAVIS-S\cite{zeng2019towards} and UHRSD\cite{xie2022pyramid} datasets. The best result is marked in \textcolor{red}{red}, while the second-best result is marked in \textcolor{green}{green}.}
\centering
\resizebox{\textwidth}{!}{  
\setlength{\arrayrulewidth}{0.4pt}
\renewcommand{\arraystretch}{1.3}
\begin{tabular}{ccc|cccc|cccc|cccc}
\toprule[2pt]

{}    & {}     & {}    & \multicolumn{4}{c|}{\emph{HRSOD-TE}}    & \multicolumn{4}{c|}{\emph{DAVIS-S}}   & \multicolumn{4}{c}{\emph{UHRSD-TE}}    \\  
\multirow{-2}{*}{\emph{Method}} & \multirow{-2}{*}{\emph{Publications}} & \multirow{-2}{*}{\emph{Year}}   & $\mathcal{M}$  \(\downarrow\)  & \(F_{\beta}^{max}\) \(\uparrow\) & \(S_m\) \(\uparrow\)  & \(E^m_{\phi}\) \(\uparrow\) & $\mathcal{M}$  \(\downarrow\)          & \(F_{\beta}^{max}\) \(\uparrow\)  & \(S_m\) \(\uparrow\)    & \(E^m_{\phi}\) \(\uparrow\)    & $\mathcal{M}$  \(\downarrow\)   & \(F_{\beta}^{max}\) \(\uparrow\)  & \(S_m\) \(\uparrow\)    & \(E^m_{\phi}\) \(\uparrow\)   \\  \midrule[1.2pt]

{HRSOD\cite{zeng2019towards}}                    & {ICCV}                           & {2019}                      & {0.030} & {0.905} & {0.896}& {0.934} & {0.026} & {0.899} & {0.876} & {0.955}       & {-}     & {-}     & {-}  & {-}   \\ 
{DRFNet\cite{zhang2021looking}}                                          & TIP                                                   & {2021}                      & {0.025} & {0.906} & {0.913}  & {-}& {0.012} & {0.904} & {0.940} & {-}  & {-}      & {-}     & {-}   & {-}  \\
HQSOD\cite{tang2021disentangled}                                           & ICCV                                                  & {2021}                      & {0.022} & {0.922} & {0.920} & {0.947} & {0.012} & {0.938} & {0.939}  & {0.974}& {0.039}  & {0.927} & {0.900} & {0.899} \\ 
DDPNet\cite{wang2022dual}                                          & AI                                                    & {2022}                      & {-}     & {0.906} & {0.901} & {-} & {-}& {-}& {-}    & {-}     & {-}        & {-}     & {-}     & {-}     \\ 

{PGNet\cite{xie2022pyramid}}                    & {CVPR}                           & {2022}                      & {0.020} & {0.937} & {0.935}  & {0.946}& {0.012} & {0.950} & {0.948} & {0.975}& {0.036} & {0.935} & \textcolor{red}{0.948} & {0.905}\\ 
{InSPyReNet\cite{kim2022revisiting}}                    & {ACCV}                           & {2022}                      & \textcolor{green}{0.014} & {0.918} & \textcolor{green}{0.956}  & {0.961}& \textcolor{green}{0.007} & {0.931} & \textcolor{green}{0.972} & {0.982}& {0.028} & {0.891} & {0.936} & {0.915}\\
RMFormer\cite{deng2023recurrent}
& {MM}                           & {2023}                      & {0.020} & {0.941} & {0.940}  & {0.949}& {0.010} & {0.955} & {0.952} & {0.978} & {0.027} & {0.949} & {0.931}  & {0.914}\\
ESNet\cite{liu2024esnet}                                           & ICML                                                  &{2024} & {0.019} & {0.937} & {0.942}  & {0.958}& {0.009} & {0.949} & {0.949} & {0.982} & {0.027} & {0.935} & {0.931} & {0.923} \\
DC-Net\cite{zhu2025dc}                                          & PR                                                   & {2025}                      & {0.068}     & {0.844} & {0.885}  & {0.837}& {0.037}& {0.789}   & {0.879}         & {0.841}     & {0.053}     & {0.844}     & {0.885}       & {0.837}     \\
\hline
SAM2\cite{ravi2024sam}                                        & ICLR                                                  &{2025} &  {0.017}      &\textcolor{green}{0.963}        &  {0.954}      &  \textcolor{green}{0.964}    &  \textcolor{green}{0.007}    &  \textcolor{green}{0.976}   & {0.963}     & {0.983}      &   \textcolor{green}{0.026} & \textcolor{green}{0.962} & {0.939} & \textcolor{red}{0.952}\\
SAM2 Adapter\cite{chen2024sam2}                                         & --                                                 &{2024} &  {0.021}      &{0.956}        &  {0.946}      &  {0.960}    &  \textcolor{green}{0.007}    &  {0.973}   & {0.965}     & {0.985}      &   {0.028} & {0.943} & {0.938} & \textcolor{green}{0.951} \\
\hline
\emph{Ours} & -- & -- & \textcolor{red}{0.012}      &\textcolor{red}{0.980}        &  \textcolor{red}{0.966}      &  \textcolor{red}{0.976}    &  \textcolor{red}{0.005}    &  \textcolor{red}{0.983}   & \textcolor{red}{0.974}     & \textcolor{red}{0.988}      &   \textcolor{red}{0.024} & \textcolor{red}{0.970} & \textcolor{green}{0.943} & \textcolor{red}{0.952}   \\

\bottomrule[2pt]
\\ 
\end{tabular}
}
\vspace{-5pt}
\end{table*}
\subsubsection{Experimental Analysis on DIS}
We compare our MGD-SAM2 with 15 previous methods on DIS5K, including $F^3$Net\cite{wei2020f3net}, GCPANet\cite{chen2020global}, PFNet\cite{mei2021camouflaged}, ISDNet\cite{guo2022isdnet}, IFA\cite{hu2022learning}, PGNet\cite{xie2022pyramid}, ISNet\cite{qin2022highly}, C2FNet\cite{chen2022camouflaged}, BCMNet\cite{cheng2023bidirectional}, UDUN\cite{pei2023unite}, FP-DIS\cite{zhou2023dichotomous}, FSANet\cite{jiang2024high}, MVANet\cite{yu2024multi}, SAM2\cite{ravi2024sam}, and SAM2 Adapter\cite{chen2024sam2}. As shown in Table \ref{tabel_dis}, our MGD-SAM2 achieves the best results across all evaluation metrics on the compared datasets. Specifically, our method outperforms the previous SOTA method, MVANet, on DIS-TE1, DIS-TE3, and DIS-TE4, and achieves comparable performance on DIS-TE2. Overall, we obtain improvements of 0.017, 0.015, 0.001, 0.007, and 0.004 in $F_{\beta}^{max}$, $F^{\omega}_{\beta}$ $E^m_{\phi}$, $S_m$, and $\mathcal{M}$, respectively. Additionally, although the fully fine-tuned SAM2 and the fine-tuned SAM2 Adapter perform well on DIS5K, they still lag significantly behind our method.
For qualitative analysis, we present the visualization results in Fig. \ref{visual_dis}. It is observed that our MGD-SAM2 generates a more accurate segmentation result in overall structure and fine-grained details compared with seven other approaches. Besides, we showcase the precision–recall and F-measure curves to evaluate the segmentation performance. As depicted in Fig. \ref{curve_dis}, the red line, representing MGD-SAM2, outperforms the other seven methods across most thresholds.

\begin{figure*}[!t]
    \centering
    \includegraphics[width=\textwidth]{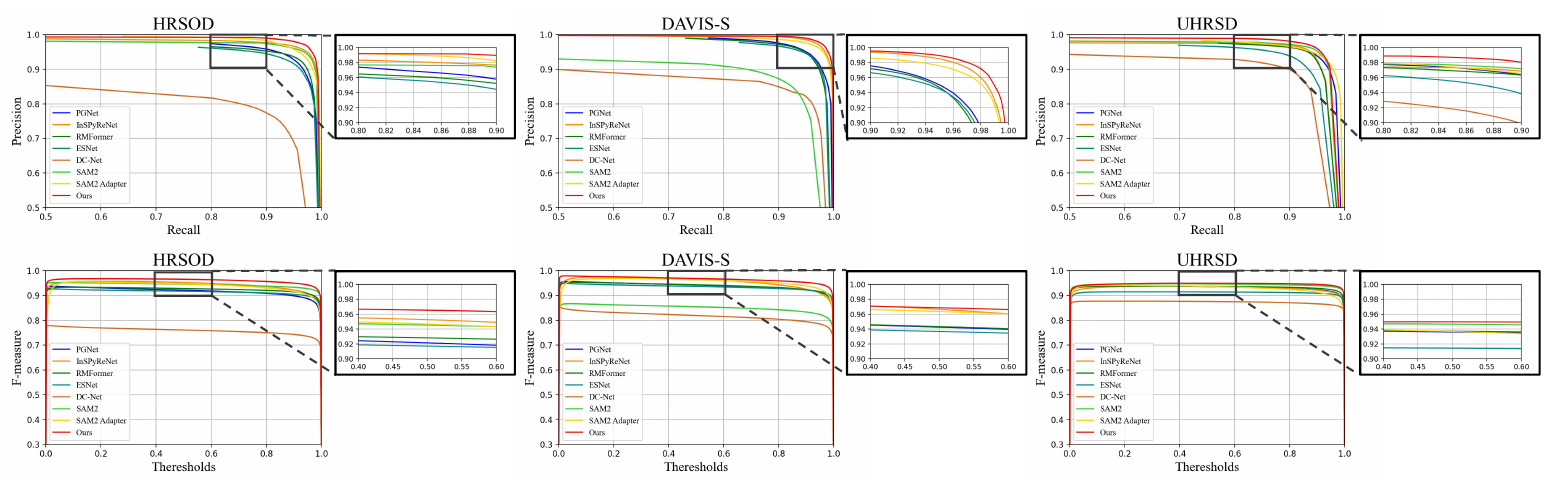}
    \caption{PR and F-measure curves of the proposed model and other existing methods on HRSOD and SOD datasets, including PGNet\cite{xie2022pyramid}, 
    InSPyReNet\cite{kim2022revisiting}, RMFormer\cite{deng2023recurrent}, ESNet\cite{liu2024esnet}, DC-Net\cite{zhu2025dc}, SAM2\cite{ravi2024sam}, and SAM2 Adapter\cite{chen2024sam2}.}
    \label{curve_sod}
    \vspace{-5pt}
\end{figure*}

\begin{table*}[htbp]
\label{table_sod}
\caption{Quantitative results on the normal-resolution DUTS\cite{wang2017learning} and HKU-IS\cite{li2015visual} datasets. The best result is marked in \textcolor{red}{red}, while the second-best result is marked in \textcolor{green}{green}.}
\centering
\resizebox{0.75\textwidth}{!}{  

\setlength{\arrayrulewidth}{0.4pt}
\renewcommand{\arraystretch}{1.3}
\begin{tabular}{ccc|cccc|cccc}
\toprule[2pt]

{}    & {}     & {}    & \multicolumn{4}{c|}{\emph{DUTS-TE}}    & \multicolumn{4}{c}{\emph{HKU-IS}}      \\  
\multirow{-2}{*}{\emph{Method}} & \multirow{-2}{*}{\emph{Publications}} & \multirow{-2}{*}{\emph{Year}}   & $\mathcal{M}$  \(\downarrow\)  & \(F_{\beta}^{max}\) \(\uparrow\) & \(S_m\) \(\uparrow\) & \(E^m_{\phi}\) \(\uparrow\) & $\mathcal{M}$  \(\downarrow\)          & \(F_{\beta}^{max}\) \(\uparrow\)  & \(S_m\) \(\uparrow\)  & \(E^m_{\phi}\) \(\uparrow\)    \\  \midrule[1.2pt]

{CPD\cite{wu2019cascaded}}                    & {CVPR}                           & {2019}                      & {0.043} & \textcolor{red}{0.972} & {0.869}  & {0.898} & {0.034} & {0.828} & {0.905} & {0.938}  \\ 
{F3Net\cite{wei2020f3net}}                  & {AAAI}                           & {2020}                      & {0.035} & {0.905} & {0.888}  & {0.920} & {0.028} & {0.943} & {0.917} & {0.952}  \\
{GAGNet-L\cite{mohammadi2020cagnet}}                    & {PR}                           & {2020}                      & {0.029} & {0.898} & {0.897}  & {0.939} & {0.024} & {0.940} & {0.923} & {0.961}  \\
{DFI\cite{liu2020dynamic}}                    & {TIP}                           & {2020}                      & {0.039} & {0.896} & {0.887}  & {0.912} & {0.031} & {0.934} & {0.920} & {0.951}  \\
{GateNet-X\cite{zhao2020suppress}}                    & {ECCV}                           & {2020}                      & {0.035} & {0.908} & {0.897}  & {0.916} & {0.029} & {0.946} & {0.925} & {0.947}  \\
{MINet\cite{pang2020multi}}                    & {CVPR}                           & {2020}                      & {0.037} & {0.884} & {0.884}  & {0.917} & {0.029} & {0.942} & {0.919} & {0.952}  \\
{LDF\cite{wei2020label}}                    & {CVPR}                           & {2020}                      & {0.034} & {0.905} & {0.892}  & {0.925} & {0.028} & {0.943} & {0.919} & {0.953}  \\
{VST\cite{liu2021visual}}                    & {ICCV}                           & {2021}                      & {0.037} & {0.895} & {0.896}  & {0.919} & {0.029} & {0.946} & {0.928} & {0.952}  \\
{ICON\cite{zhuge2022salient}}                    & {TPAMI}                           & {2022}                      & {0.025} & {0.924} & {0.917}  & {0.954} & {0.022} & {0.954} & {0.935} & {0.968}  \\
{InSPyReNet\cite{kim2022revisiting}}                    & {ACCV}                           & {2022}                      & {0.024} & {0.860} & {0.934}  & {0.928}& {0.023} & {0.923} & {0.934} & {0.964} \\
{BBRF\cite{ma2023boosting}}                    & {TIP}                           & {2023}                      & {0.025} & {0.911} & {0.909}  & {0.949} & {0.020} & {0.949} & {0.932} & \textcolor{red}{0.969}  \\
MENet\cite{wang2023pixels}      & CVPR                                      & {2023}                      & {0.028}     & {0.918} & {0.905} & {0.938} & {0.023}& {0.951}   & {0.927} & {0.960}             \\
{ADMNet\cite{zhou2024admnet}}                    & {TMM}                           & {2024}                      & {0.052} & {0.813} & {0.849}  & {0.893} & {0.036} & {0.906} & {0.901} & {0.946}  \\
ESNet\cite{liu2024esnet}                                           & ICML                                                  &{2024} &  \textcolor{green}{0.022} & {0.921} & {0.924}  & {0.934}& {0.019} & {0.951} & {0.939} & {0.965}\\
{DC-Net\cite{zhu2025dc}}                    & {PR}                           & {2025}                      & {0.023} & {0.932} & {0.925}  &  \textcolor{green}{0.952} & {0.021} & {0.957} & {0.941} & {0.966}  \\
\hline
{SAM\cite{kirillov2023segment}}                    & {ICCV}                           & {2023}                      & {0.030} & {0.921} & {0.909}  & {0.937} & {0.022} & {0.956} & {0.935} & {0.965}  \\
{MDSAM\cite{gao2024multi}}                    & {MM}                           & {2024}                      & {0.024} & {0.937} & {0.920}  & {0.949} &  \textcolor{green}{0.019} & {0.963} & {0.941} & \textcolor{red}{0.969}  \\
SAM2\cite{ravi2024sam}                                        & ICLR                                                  &{2025} &  {0.024}      &{0.943}        &   \textcolor{green}{0.927}      &  {0.951}    &   \textcolor{green}{0.019}      & \textcolor{green}{0.965}        &   \textcolor{green}{0.943}    &   \textcolor{green}{0.968}    \\
SAM2 Adapter\cite{chen2024sam2}                                         & --                                                &{2024} &  {0.029}      &{0.922}        &  {0.921}      &  {0.942}    &  {0.021}    &  {0.952}   & {0.942}     & {0.967}       \\
\hline
\emph{Ours}& -- & -- &  \textcolor{red}{0.018}      & \textcolor{green}{0.963}        &  \textcolor{red}{0.941}      &  \textcolor{red}{0.963}    &  \textcolor{red}{0.018}    &  \textcolor{red}{0.970}   & \textcolor{red}{0.946}     & \textcolor{red}{0.969}    \\

\bottomrule[2pt]
\\ 
\end{tabular}
}
\vspace{-10pt}
\end{table*}

\subsubsection{Experimental Analysis on HRSOD}

We compare the proposed MGD-SAM2 with 11 previous methods (HRSOD\cite{zeng2019towards}, DRFNet\cite{zhang2021looking}, HQSOD\cite{tang2021disentangled}, DDPNet\cite{wang2022dual}, PGNet\cite{xie2022pyramid}, InSPyReNet\cite{kim2022revisiting}, 
RMFormer\cite{deng2023recurrent}, ESNet\cite{liu2024esnet}, DC-Net\cite{zhu2025dc},
SAM2\cite{ravi2024sam}, and SAM2 Adapter\cite{chen2024sam2}) on three HRSOD datasets. As illustrated in Table \ref{table_hrsod}, our MGD-SAM2 achieves the new SOTA results across nearly all evaluation metrics on the three compared datasets, except for $S_m$ on UHRSD-TE. Concretely, full fine-tuned SAM2 achieves better performance than other tailored models for HRSOD. Besides, SAM2 Adapter achieves promising results with only training the inserted Adapter and SAM2 Decoder. However, by leveraging SAM2's rich visual prior and multi-view complementarity, our method surpasses SAM2 by a significant margin with a few trainable parameters (See trainable parameters comparison of SAM2 and our MGD-SAM2 on Table \ref{ablation_conponets}). We also provide a visual comparison of our MGD-SAM2 and other methods on HRSOD, DAVIS-S, and UHRSD. As shown in the first six lines of Fig. \ref{visual_sod}, our MGD-SAM2 consistently achieves better segmentation results in global structure and local details across all three HRSOD datasets. These visual results demonstrate that our framework significantly enhances the fine-grained segmentation on HRCS tasks, yielding improved prediction masks by leveraging SAM2's rich visual representations and multi-view interaction. We also present the precision–recall and F-measure curves in the first three columns of Fig. \ref{curve_sod}, further demonstrating the superiority of our approach.

\subsubsection{Experimental Analysis on SOD}

To evaluate the generalization ability of our MGD-SAM2, we further compare the proposed approach with 18 previous methods (CPD\cite{wu2019cascaded}, F3Net\cite{wei2020f3net}, GAGNet-L\cite{mohammadi2020cagnet}, DFI\cite{liu2020dynamic}, GateNet-X\cite{zhao2020suppress}, MINet\cite{pang2020multi}, LDF\cite{wei2020label}, VST\cite{liu2021visual}, ICON\cite{zhuge2022salient}, InSPyReNet\cite{kim2022revisiting}, BBRF\cite{ma2023boosting}, MENet\cite{wang2023pixels}, ADMNet\cite{zhou2024admnet}, ESNet\cite{liu2024esnet}, DC-Net\cite{zhu2025dc}, SAM\cite{kirillov2023segment}, MDSAM\cite{gao2024multi}, SAM2\cite{ravi2024sam}, and SAM2 Adapter\cite{chen2024sam2}) on two normal-resolution datasets. Consistently, our MGD-SAM2 obtains new SOTA outcomes in both DUTS-TE and HKU-IS. As shown in Table \ref{table_sod}, MDSAM outperforms SAM in the SOD task by fine-tuning it with multi-level and multi-stage features. Additionally, SAM2's superior zero-shot segmentation capability also enables better results than SAM. However, the above SAM-based methods only achieve comparable or worse performance than tailored SOD methods (i.e. ESNet and DC-Net). In contrast, by leveraging multi-view complementarity and detail enhancement, our method significantly improves the segmentation accuracy on the basis of SAM2. As illustrated in the last four lines of Fig. \ref{visual_sod}, our method can identify accurate salient objects with less mis-segmentation of background regions on the two normal-resolution datasets. Besides, the precision–recall and F-measure curves presented in the last two columns of Fig. \ref{curve_dis} verify the superiority of our approach.

\subsection{Ablation Study}
\begin{table}[htbp]
\centering
\caption{Ablation studies of the MPAdapter on DIS-VD. The best result is marked in \textbf{bold}}
\label{ablation_mpadapter}
\resizebox{0.5\textwidth}{!}{  
\setlength{\arrayrulewidth}{0.4pt}
\renewcommand{\arraystretch}{1.3}
\begin{tabular}{c|cc|cccc} 
\toprule[2pt]
         {}&  Tuned &Inserted &\multicolumn{4}{c}{{DIS-VD}} \\
         \multirow{-2}{*}{\emph{Method}} &Params(M) &Params(M)
         &$\mathcal{M}$\(\downarrow\)
         &  \(F_{\beta}^{max} \uparrow\)&\(S_m \uparrow\)
         & \(E^m_{\phi} \uparrow\)  \\ \hline 
           (a) Full fine-tuning &  72.910  &  -&  0.038& 0.899& 0.888 & 0.918\\ 
         (b) SAM2 Adapter \cite{chen2024sam2}&  4.315 &  0.081& 0.045& 0.890&0.881&0.912\\
         (c) Adapter \cite{houlsby2019parameter}&  4.678 &  0.441& 0.041& 0.907&0.890&0.923\\
         (d) LoRA \cite{hu2022lora} &  4.315&  0.319& 0.040& 0.906&0.889&0.919\\ 
         (e) LMSA \cite{gao2024multi} &  6.665&  2.430& 0.040&0.903& 0.891&0.921\\
\hline
(f) MPAdapter &4.724 &  0.489&  \textbf{0.037} &  \textbf{0.913}& \textbf{0.899}&\textbf{0.926}\\ 
\bottomrule[2pt]
\end{tabular}
}
\vspace{-5pt}
\end{table}

\begin{table}[htbp]
\caption{Ablation experiments of each component on DIS-VD. The best result is marked in \textbf{bold}.}
\centering
\label{ablation_conponets}
\resizebox{0.5\textwidth}{!}{  
\setlength{\arrayrulewidth}{0.4pt}
\renewcommand{\arraystretch}{1.4}
\begin{tabular}{c|cc|cccc} 
\toprule[2pt]
         {}&  Tuned &Inserted &\multicolumn{4}{c}{{DIS-VD}} \\
         \multirow{-2}{*}{\emph{Method}} &Params(M) &Params(M)
         &$\mathcal{M}$\(\downarrow\)
         &  \(F_{\beta}^{max} \uparrow\)&\(S_m \uparrow\)
         & \(E^m_{\phi} \uparrow\)  \\ \hline 
           (a) Full fine-tuning &  72.910  &  -&  0.038& 0.899& 0.888 & 0.918\\ 
         (b) MPAdapter &  4.724 &  0.489&  0.037 &  0.913& 0.899&0.926\\
         (c) MPAdapter+MCEM  &  5.910 &  1.676&  0.035& 0.916& 0.901&0.930\\ 
         (d) MPAdapter+MCEM+HMIM &  6.578 &  2.344&  0.035& 0.918& 0.906&\textbf{0.933}\\
(e) MPAdapter+MCEM+HMIM+DRM &6.637 &2.403 & \textbf{0.034}& \textbf{0.928}& \textbf{0.908}&\textbf{0.933}\\ 
\bottomrule[2pt]
\end{tabular}
}
\vspace{-5pt}
\end{table}

To verify the effectiveness of MGD-SAM2's components modules, we conduct ablation studies in this section. If not specified, all the experiments are with a multi-view input.

\subsubsection{Effectiveness of MPAdapter} In Table \ref{ablation_mpadapter}, we compare our MPAdapter with full fine-tuning and four parameter-efficient fine-tuning methods: SAM2 Adapter \cite{chen2024sam2}, Adapter \cite{houlsby2019parameter}, LoRA \cite{hu2022lora}, LMSA in MDSAM \cite{gao2024multi}. We only change the backbone of SAM2 and keep the neck and decoder of SAM2 trainable for all methods. Specifically, the introduction of fine-tuning methods significantly reduces the trainable parameters, while achieving comparable performance. However, our MPAdapter enables more precise segmentation with similar parameters compared to those fine-tuning methods.

\subsubsection{Effectiveness of MCEM} Table \ref{ablation_conponets} (c) illustrates the quantitative results with MCEM. It is observed that the model achieves an improvement in all metrics with MCEM. These results demonstrate that although the lightweight MPAdapter improves the perception of global and local features, the extracted deep feature $F_{16}$ can be further enhanced via the attention mechanism. Besides, we illustrate the visual comparison of feature maps to verify the effectiveness of our MCEM. As shown in the feature maps of $F_{16}$ and $E_{16}$ in Fig. \ref{feature}, MCEM improves the foreground object integrity and morphological structure of the feature maps.

\subsubsection{Effectiveness of HMIM} 
\begin{figure*}[htbp]
    \centering
    \includegraphics[width=\textwidth]{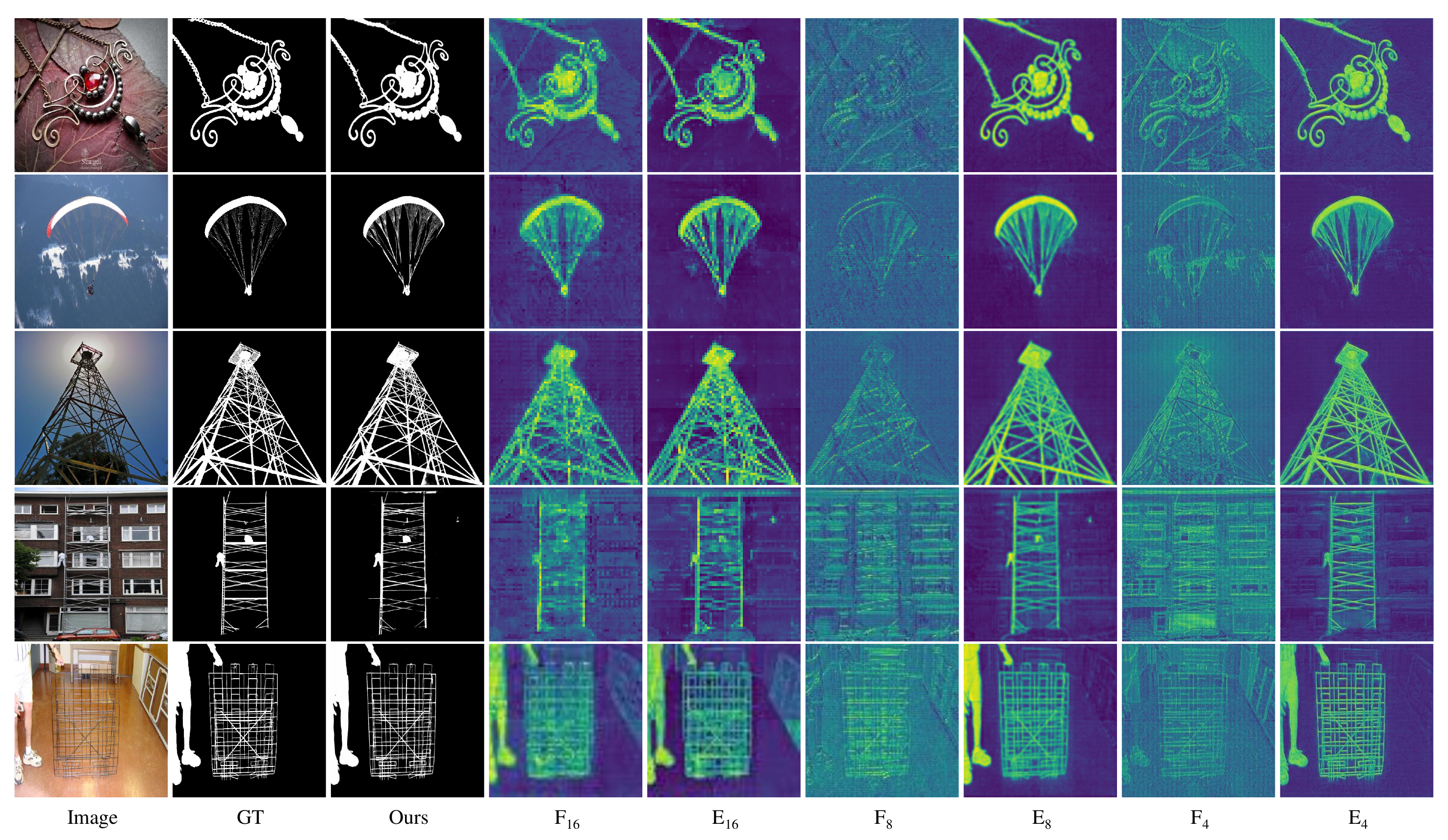}
    \caption{ The visualization of feature maps and segmentation results from our MGD-SAM2 on the DIS5K-VD. $\{F_4,F_8\}$ denotes the shallow features from the MPAdapter-assisted SAM2 encoder. $F_{16}$ is the output deep feature of Two Way
 Transformer. $\{E_4,E_8,E_{16}\}$ represents the enhanced multi-stage features from the MCEM and HMIM. $\{4,8,16\}$ is the downsampling stride. All the feature maps use the merged local features.}
    \label{feature}
\end{figure*}

Table \ref{ablation_conponets} (d) illustrates the results with HMIM. When incorporating HMIM to leverage multi-scale and multi-view features, the model results in a notable improvement in  $F_{\beta}^{max}$, $S_m$ and
$E^m_{\phi}$, which verifies the effectiveness of HMIM in enhancing local-global representation and improving segmentation accuracy. We also present a visualization of feature maps in Fig. \ref{feature} to evaluate the effectiveness of HMIM. As shown in the comparison of $\{F_4,F_8\}$ and $\{E_4,E_8\}$, by leveraging multi-stage and multi-view features, HMIM obtains features that focus on the foreground object with fine-grained details.

\subsubsection{Effectiveness of DRM}
Table \ref{ablation_conponets} (e) illustrates the quantitative results with DRM. By employing DRM for gradually restored high-resolution results, the model gains an improvement in $\mathcal{M}$, $F_{\beta}^{max}$ and $S_m$, which demonstrates the effectiveness of DRM in generating more fine-grained mask prediction for HRCS task.

\subsection{Zero-shot Analysis}
Considering that SAMs\cite{kirillov2023segment,ravi2024sam} have demonstrated strong zero-shot segmentation capability in natural scenes, we conduct experiments in Table \ref{zero-shot} to evaluate their performance on HRCS. Both SAM and SAM2 are evaluated under two settings: automatic mode (Auto) and ground truth bounding box (Box) mode. Besides, we also test the zero-shot ability of SAM-HQ\cite{ke2023segment}, which enables SAM to generate high-quality masks. In the automatic setting, SAM, SAM-HQ, and SAM2 all achieve pool performance. SAM2 achieves an $\mathcal{M}$ of 0.156 on DIS-VD compared to 0.283 for SAM-HQ and 0.258 for SAM. With ground truth boxes as prompts, the experiment results in the box setting significantly surpass those in the automatic setting. Concretely, SAM demonstrates significant gains across all metrics, with $\mathcal{M}$, $F_{\beta}^{max}$, $ S_m$, $E^m_{\phi}$ values of 0.150, 0.671, 0.681 and 0.774. Both SAM-HQ and SAM2 show better segmentation accuracy than SAM across all metrics. By retraining with high-quality samples, SAM-HQ even outperforms SAM2. However, our method demonstrates a significant improvement over SAM-HQ, highlighting the limitations of the aforementioned zero-shot methods in segmenting detailed structures.
\begin{table}[htbp]
\caption{Zero-shot performance comparisons on DIS-VD. The best result is marked in \textbf{bold}}
\centering
\resizebox{0.4\textwidth}{!}{  
\setlength{\arrayrulewidth}{0.4pt}
\renewcommand{\arraystretch}{1.3}
\begin{tabular}{c|cccc}

\toprule[2pt]
         {} &\multicolumn{4}{c}{{DIS-VD}} \\
         \multirow{-2}{*}{\emph{Method}} 
         &$\mathcal{M}$\(\downarrow\)
         &  \(F_{\beta}^{max} \uparrow\)&\(S_m \uparrow\)
         & \(E^m_{\phi} \uparrow\)  \\ \hline 
           (a) SAM (Auto) \cite{kirillov2023segment} &  0.258& 0.215& 0.398 & 0.392\\ 
         (b) SAM-HQ (Auto) \cite{ke2023segment}& 0.283 & 0.214 & 0.374 & 0.538 \\
         (c) SAM2 (Auto) \cite{ravi2024sam}& 0.156& 0.428&0.515&0.477\\
         \hline
         (d) SAM (Box) \cite{kirillov2023segment}& 0.150& 0.671&0.681&0.774\\ 
         (e) SAM-HQ (Box) \cite{ke2023segment} & 0.053 & 0.832 & 0.839 & 0.909 \\
         (f) SAM2 (Box) \cite{ravi2024sam} & 0.104&0.765& 0.766&0.840\\
\hline
(g) Ours &\textbf{0.034}& \textbf{0.928}& \textbf{0.908}&\textbf{0.933}\\ 
\bottomrule[2pt]
\end{tabular}
}
\label{zero-shot}
\end{table}
\section{Conclusion and Future Work}
In this paper, we promote the exploration of SAM2 with multi-view complementary for HRCS task, which effectively leverages the rich visual priors of SAM2 and local-global interaction to achieve fine-grained segmentation. Specifically, we first introduce the MPAdapter to adapt SAM to HHRCS and enable the model to extract better local details and global semantics. Then, we propose the MCEM and HMIM to further exploit local texture and global context by aggregating multi-view and multi-scale features. Finally, we devise the DRM to generate the gradually restored high-resolution mask prediction. Experimental results demonstrate that our MGD-SAM2 achieves superior performance and strong generalization on multiple high-resolution and normal-resolution datasets, significantly outperforming other SOTA methods. In our future work, we will focus on lightweight design and interactive segmentation, leveraging the integration of multi-view interaction with SAM2 in a more efficient and user-friendly manner.

\bibliography{reference}
\bibliographystyle{IEEEtran}

\end{document}